\documentclass[10pt]{article}
\pdfoutput=1
\usepackage[a4paper]{geometry}
\usepackage{amsmath}
\usepackage{amssymb}
\usepackage{graphicx}

\newcommand{\address}[1]{
	\par {\raggedright #1
	\vspace{1.4em}
	\noindent\par}
}

\DeclareMathOperator{\sign}{sign}

\begin{document}
\title{Differential equations as models of deep neural networks}
\author{Julius Ruseckas}
\maketitle

\address{Baltic Institute of Advanced Technology (BPTI), Pilies 16-8, LT-01403,
Vilnius, Lithuania\\
\texttt{julius.ruseckas@bpti.lt}}

\begin{abstract}
In this work we systematically analyze general properties of differential
equations used as machine learning models. We demonstrate that the
gradient of the loss function with respect to to the hidden state
can be considered as a generalized momentum conjugate to the hidden
state, allowing application of the tools of classical mechanics. In
addition, we show that not only residual networks, but also feedforward
neural networks with small nonlinearities and the weights matrices
deviating only slightly from identity matrices can be related to the
differential equations. We propose a differential equation describing
such networks and investigate its properties.
\end{abstract}

\section{Introduction}

Deep learning is a form of machine learning that uses neural networks
with many hidden layers \cite{Bengio2009,LeCun2015}. Deep learning
models have dramatically improved speech recognition, visual object
recognition, object detection and many other domains \cite{LeCun2015}.
Since the number of layers in deep neural networks become large, it
is possible to consider the layer number as a continuous variable
\cite{Sonoda2017} and represent the neural network by an differential
equation.

The connection between the neural networks and differential equations
first appeared with an additive model for continuous time recurrent
neural networks, described by the differential equations \cite{Grossberg2013}
\begin{equation}
\tau_{i}\frac{dx_{i}}{dt}=-x_{i}+
\sum_{j=1}^{n}w_{j,i}\sigma(x_{j}-\theta_{j})+I_{i}(t)\,.
\end{equation}
Cohen and Grossberg applied the Liapunov method to prove that global
limits exist in a class of dynamical systems with symmetric interaction
coefficients, which includes the additive and shunting models \cite{Cohen1983}.
Hopfield\textquoteright s work \cite{Hopfield1984} pioneered the
analog computation of continuous time recurrent neural networks instead
of digital computation using complex numerical algorithms on a digital
computer. A Hopfield network has a quadratic form as an Lyapunov function
for the activity dynamics. As a consequence, the state of the network
evolves to a final state that is a minimum of the Lyapunov function
when started in any initial state \cite{Hopfield2007}. Continuous
time recurrent neural networks have been applied to address vision
\cite{Harvey1994}, cognitive behavior \cite{Beer1997}, and cooperation
\cite{Quinn2001}.

New relations between neural network and differential equation appeared
after proposal of residual networks (ResNets) \cite{He2016}. Forward
propagation of discrete vector $\boldsymbol{x}\in\mathbb{R}^{n}$
through residual network can be written as
\begin{equation}
\boldsymbol{x}^{(l+1)}=\boldsymbol{x}^{(l)}+
F(\boldsymbol{x}^{(l)},\boldsymbol{q}^{(l)})\,,\label{eq:resnet}
\end{equation}
where $\boldsymbol{q}^{(l)}$ are the parameters that need to be determined
by the training of the neural network. In \cite{Liao2016} a connection
of residual network with recurrent neural network, which is known
as an approximation of a dynamical system, has been made. Furthermore,
equation (\ref{eq:resnet}) can be seen as an Euler discretization
\begin{equation}
\frac{1}{\Delta t}(\boldsymbol{x}^{(l+1)}-\boldsymbol{x}^{(l)})=
F(\boldsymbol{x}^{(l)},\boldsymbol{q}^{(l)})\,,\qquad\Delta t=1
\end{equation}
 of an ordinary differential equation (ODE)
\begin{equation}
\frac{d}{dt}\boldsymbol{x}(t)=F(\boldsymbol{x}(t),\boldsymbol{q}(t))\,.
\end{equation}
Such a link between residual neural networks and nonlinear differential
equations has been first proposed in \cite{E2017,Haber2017}. This
interpretation of neural network as a discretization of differential
equation has been applied to analyze stability properties of deep
learning \cite{Haber2018} and to derive reversible neural network
architectures \cite{Chang2017}. The view of ResNets as an Euler discretization
of ODEs allowed to construct invertible ResNets (i-ResNets) \cite{Behrmann2018}.
Related to this interpretation is a view of a finite deep neural network
as a broken line approximation of a continuous trajectory \cite{Sonoda2017}
and the formulation of the training process of ResNet as an control
problem of a transport equation \cite{Li2017}. Subsequently it has
been demonstrated \cite{Lu2017} that many networks, such as ResNet,
PolyNet, FractalNet and RevNet, can be interpreted as different numerical
discretizations of differential equations and linear multi-step architecture
has been proposed. In addition to interpreting the layer depth as
a continuous variable, the input to the neural network can also be
represented as being continuous. Such a view has been taken in \cite{Ruthotto2018},
where residual convolutional neural networks have been interpreted
as a discretization of a space-time partial differential equation
(PDE). In \cite{Long2017,Long2018} a new feed-forward deep network,
PDE-Net, has been proposed to predict dynamics of complex systems
and to uncover the underlying hidden PDE models. Further connection
between RNNs and ODEs has been revealed in \cite{Niu2019}. It has
been shown that the temporal dynamics of RNN architectures can be
represented by a specific numerical integration method for ODEs of
a given order.

Using differential equations a family of models called Neural ODEs has been
proposed in \cite{Chen2018}. The output of the network is computed using a
black-box differential equation solver and the gradients are computed by
solving a second, augmented ODE backwards in time \cite{Chen2018}. These models
have shown great promise on a number of tasks including modeling continuous
time data and building normalizing flows with low computational cost.  In
particular, ODEs have been applied to model time series with non-uniform
intervals by generalizing state transitions in RNNs to continuous-time dynamics
\cite{Rubanova2019}.  However, neural ODEs learn representations that preserve
the topology of the input space, therefore there exist functions that neural
ODEs cannot represent \cite{Dupont2019}. Usual ResNets, being a discretization
of the ODE, allow the trajectories to make discrete jumps to cross each other
and do not have this limitation.  To address the limitations of neural
differential equations, Augmented Neural ODEs are introduced by adding
additional dimensions \cite{Dupont2019}.

In this work we first systematically analyze general properties of
differential equations used as machine learning models. We demonstrate
that the gradient of the loss function with respect to to the hidden
state can be considered as a generalized momentum conjugate to the
hidden state, allowing application of the tools of classical mechanics.
In addition, we show that not only residual networks, but also feedforward
neural networks with small nonlinearities and the weights matrices
deviating only slightly from identity matrices can be related to the
differential equations. We propose a differential equation describing
such networks and investigate its properties.

This paper is organized as follows. In section~\ref{sec:odes} we
investigate general properties of first-order nonlinear differential
equations as models in machine learning. In section~\ref{sec:feedforward-ode}
we propose a differential equation describing feedforward neural networks
and in section~\ref{sec:properties} we analyze its properties. In
section~\ref{sec:examples} we apply the proposed differential equation
for a couple of simple problems. Finally, section~\ref{sec:conclusions}
summarizes our results.

\section{Ordinary differential equations as machine learning models\label{sec:odes}}

In this section we investigate nonlinear ordinary differential equations
used as machine learning models. We assume the differential equations
of the form
\begin{equation}
\frac{d}{dt}\boldsymbol{x}(t)=
\boldsymbol{F}(\boldsymbol{x}(t),\boldsymbol{q}(t))\,,\label{eq:diff-general}
\end{equation}
where $\boldsymbol{F}$ are nonlinear functions and $\boldsymbol{q}(t)$
are arbitrary parameters that are functions of $t$. Equation (\ref{eq:diff-general})
is more general than the differential equations considered in \cite{Chen2018},
where the parameters do not depend on $t$. Starting from the input
$\boldsymbol{x}(0)$, the output $\boldsymbol{x}(T)$ is the solution
at $t=T$ to the initial value problem of the differential equation
(\ref{eq:diff-general}). The output $\boldsymbol{x}(T)$ corresponds
to the features learned by the model. The parameters $\boldsymbol{q}(t)$
are adjusted such that equation (\ref{eq:diff-general}) transforms
the input $\boldsymbol{x}(0)$ to a $\boldsymbol{y}$ which is close
to true label $\boldsymbol{y}_{\mathrm{true}}$.

\subsection{Backward propagation for differential equations}

Let us consider a loss function $\mathcal{L}(\boldsymbol{x}(T))$ that is a
function of a final output $\boldsymbol{x}(T)$. The minimum of the loss can be
found by applying the gradient descent method.  When $t$ is a continuous
variable, the loss function $\mathcal{L}(\boldsymbol{x}(T))$ is a functional of
functions $\boldsymbol{q}(t)$ and should be minimized with the constraint
$\dot{\boldsymbol{x}}-\boldsymbol{F}(\boldsymbol{x},\boldsymbol{q})=0$, where
for the brevity we use the notation $\dot{\boldsymbol{x}}\equiv
d\boldsymbol{x}/dt$.  A general method how to find such a minimum by
introducing backpropagating dynamics is given by Pontryagin's minimum principle
\cite{Pontryagin2018}. In this section we briefly sketch the derivation of the
backpropation corresponding to equation (\ref{eq:diff-general}).  In order to
derive update rule for for the parameters $\boldsymbol{q}(t)$ one needs to use
calculus of variations and compute the functional derivatives
\cite{Giaquinta2006}. For the minimization with constraints we can employ the
method of Lagrange multipliers. Thus we will consider the functional
\begin{equation}
S=\mathcal{L}(\boldsymbol{x}(T))+
\int_{0}^{T}\boldsymbol{a}(t)[\boldsymbol{F}(\boldsymbol{x},\boldsymbol{q})
-\dot{\boldsymbol{x}}]\,dt\,,\label{eq:lagrangian-1}
\end{equation}
where $\boldsymbol{a}(t)$ is a vector of Lagrange multipliers. Since
the constraint $\dot{\boldsymbol{x}}-\boldsymbol{F}(\boldsymbol{x},\boldsymbol{q})=0$
is satisfied by construction, we are free to set the values of $\boldsymbol{a}$.
We need to calculate the functional derivative $\delta\mathcal{L}/\delta\boldsymbol{q}(t)
=\delta S/\delta\boldsymbol{q}(t)$.
Using equation (\ref{eq:lagrangian-1}) we obtain
\begin{equation}
\frac{\delta S}{\delta\boldsymbol{q}(t)}=
\frac{\partial\mathcal{L}}{\partial\boldsymbol{x}(T)}
\frac{\delta\boldsymbol{x}(T)}{\delta\boldsymbol{q}(t)}
+\int_{0}^{T}\boldsymbol{a}(t')\left[\frac{\partial}{\partial\boldsymbol{x}(t')}
\boldsymbol{F}(\boldsymbol{x}(t'),\boldsymbol{q}(t'))
\frac{\delta\boldsymbol{x}(t')}{\delta\boldsymbol{q}(t)}+
\frac{\partial}{\partial\boldsymbol{q}(t')}
\boldsymbol{F}(\boldsymbol{x}(t'),\boldsymbol{q}(t'))\delta(t-t')
-\frac{\delta\dot{\boldsymbol{x}}(t')}{\delta\boldsymbol{q}(t)}\right]dt'\,.\label{eq:func-deriv-1}
\end{equation}
The last term we integrate by parts:
\begin{equation}
\int_{0}^{T}\boldsymbol{a}(t')
\frac{\delta\dot{\boldsymbol{x}}(t')}{\delta\boldsymbol{q}(t)}\,dt'=
\left.\boldsymbol{a}\frac{\delta\boldsymbol{x}(t')}{\delta\boldsymbol{q}(t)}\right|_{t'=0}^{t'=T}
  -\int_{0}^{T}\dot{\boldsymbol{a}}(t')\frac{\delta\boldsymbol{x}(t')}{\delta\boldsymbol{q}(t)}\,dt'\,.
\end{equation}
The initial condition $\boldsymbol{x}(0)$ does not depend on the
parameters $\boldsymbol{q}$, therefore $\delta\boldsymbol{x}(0)/\delta\boldsymbol{q}(t)=0$.
Inserting into equation (\ref{eq:func-deriv-1}) we have
\begin{equation}
\frac{\delta S}{\delta\boldsymbol{q}(t)}=
\left(\frac{\partial\mathcal{L}}{\partial\boldsymbol{x}(T)}-
\boldsymbol{a}(T)\right)\frac{\delta\boldsymbol{x}(T)}{\delta\boldsymbol{q}(t)}+
\int_{0}^{T}\left(\dot{\boldsymbol{a}}(t')+
\boldsymbol{a}(t')\frac{\partial}{\partial\boldsymbol{x}(t')}
\boldsymbol{F}(\boldsymbol{x}(t'),\boldsymbol{q}(t'))\right)
\frac{\delta\boldsymbol{x}(t')}{\delta\boldsymbol{q}(t)}\,dt'+
\boldsymbol{a}(t)\frac{\partial}{\partial\boldsymbol{q}(t)}
\boldsymbol{F}(\boldsymbol{x}(t),\boldsymbol{q}(t))\,.\label{eq:func-deriv-2}
\end{equation}
Since $\boldsymbol{a}(t)$ can be arbitrary, we can set the coefficients
in equation (\ref{eq:func-deriv-2}) to be zero:
\begin{equation}
\boldsymbol{a}(T)=\frac{\partial\mathcal{L}}{\partial\boldsymbol{x}(T)}
\end{equation}
and
\begin{equation}
-\frac{d}{dt}\boldsymbol{a}(t)=\boldsymbol{a}(t)
\frac{\partial}{\partial\boldsymbol{x}(t)}
\boldsymbol{F}(\boldsymbol{x}(t),\boldsymbol{q}(t))\,.\label{eq:backward-general}
\end{equation}
Equation (\ref{eq:backward-general}), representing the differential
equation for backward propagation, is adjoint to the equation (\ref{eq:diff-general})
\cite{Pontryagin2018}. It is a continuous-time analog to the usual
backporpagation algorithm \cite{Rumelhart1986}. In contrast to the
non-linear equation (\ref{eq:diff-general}) for the forward propagation,
equation (\ref{eq:backward-general}) for the backward propagation
of the gradient is linear. The remaining term in equation (\ref{eq:func-deriv-2})
gives
\begin{equation}
\frac{\delta\mathcal{L}}{\delta\boldsymbol{q}(t)}=
\boldsymbol{a}(t)\frac{\partial}{\partial\boldsymbol{q}(t)}
\boldsymbol{F}(\boldsymbol{x}(t),\boldsymbol{q}(t))\,.\label{eq:func-deriv-3}
\end{equation}
An alternative derivation of equation (\ref{eq:backward-general})
that does not involve variational calculus is presented in the Appendix~\ref{sec:appendix}.

If the functions $\boldsymbol{q}(t)$ depend on parameters $\boldsymbol{\theta}$
that do not depend on $t$, $\boldsymbol{q}(t)=\boldsymbol{q}(\boldsymbol{\theta},t)$,
then the gradient of the loss can be calculated as
\begin{equation}
\frac{\partial\mathcal{L}}{\partial\boldsymbol{\theta}}=
\int_{0}^{T}\frac{\delta\mathcal{L}}{\delta\boldsymbol{q}(t)}
\frac{\partial\boldsymbol{q}(t)}{\partial\boldsymbol{\theta}}\,dt\,.
\end{equation}
Using equation (\ref{eq:func-deriv-3}) we get
\begin{equation}
\frac{\partial\mathcal{L}}{\partial\boldsymbol{\theta}}=
\int_{0}^{T}\boldsymbol{a}(t)\frac{\partial}{\partial\boldsymbol{q}(t)}
\boldsymbol{F}(\boldsymbol{x}(t),\boldsymbol{q}(t))
\frac{\partial\boldsymbol{q}(t)}{\partial\boldsymbol{\theta}}\,dt\,.\label{eq:gradient-param}
\end{equation}
In particular, if the parameters $\boldsymbol{q}$ do not depend on
$t$ then we can take $\boldsymbol{\theta}=\boldsymbol{q}$ and from
equation (\ref{eq:gradient-param}) obtain
\begin{equation}
\frac{\partial\mathcal{L}}{\partial\boldsymbol{q}}=
\int_{0}^{T}\boldsymbol{a}(t)\frac{\partial}{\partial\boldsymbol{q}}
\boldsymbol{F}(\boldsymbol{x}(t),\boldsymbol{q})\,dt\,.\label{eq:grdient-fixed-param}
\end{equation}
The total time $T$ can also be interpreted as one of the parameters
of the model. The gradient of the loss $\mathcal{L}(\boldsymbol{x}(T))$
with respect to the total time $T$ is
\begin{equation}
\frac{d}{dT}\mathcal{L}(\boldsymbol{x}(T))=
\frac{\partial\mathcal{L}}{\partial\boldsymbol{x}(T)}
\frac{d\boldsymbol{x}(T)}{dT}=\boldsymbol{a}(T)\boldsymbol{F}(\boldsymbol{x}(T),\boldsymbol{q}(T))\,.
\end{equation}

\subsection{Euler-Lagrange and Hamiltonian form of the equations}

Equations (\ref{eq:diff-general}) and (\ref{eq:backward-general})
can be written in the form of the Euler-Lagrange equations
\begin{align}
\frac{\partial L}{\partial\boldsymbol{a}}-
\frac{d}{dt}\frac{\partial L}{\partial\dot{\boldsymbol{a}}} & =0\,,\\
\frac{\partial L}{\partial\boldsymbol{x}}-
\frac{d}{dt}\frac{\partial L}{\partial\dot{\boldsymbol{x}}} & =0
\end{align}
with the Lagrangian
\begin{equation}
L(\boldsymbol{x},\boldsymbol{a},\dot{\boldsymbol{x}},t)=
\boldsymbol{a}\dot{\boldsymbol{x}}-\boldsymbol{a}
\boldsymbol{F}(\boldsymbol{x},\boldsymbol{q}(t))\,.\label{eq:lagrangian}
\end{equation}
Here $\boldsymbol{x}$ and $\boldsymbol{a}$ play the role of generalized
coordinates. Lagrangian of this type for differential equations considered
together with adjoint equations has been proposed in \cite{Ibragimov2006}.
This Lagrangian (\ref{eq:lagrangian}) appears in equation (\ref{eq:lagrangian-1}).
The Lagrangian (\ref{eq:lagrangian}) yields the generalized momentum
\begin{equation}
\frac{\partial L}{\partial\dot{\boldsymbol{x}}}=\boldsymbol{a}
\end{equation}
leading to the corresponding Hamiltonian
\begin{equation}
H(\boldsymbol{x},\boldsymbol{a},t)=\boldsymbol{a}\dot{\boldsymbol{x}}-L=
\boldsymbol{a}\boldsymbol{F}(\boldsymbol{x},\boldsymbol{q}(t))\,.\label{eq:hmiltonian}
\end{equation}
Then the equations (\ref{eq:diff-general}) and (\ref{eq:backward-general})
take the form of the Hamilton equations
\begin{align}
\frac{d\boldsymbol{x}}{dt} & =\frac{\partial H}{\partial\boldsymbol{a}}\,,\\
\frac{d\boldsymbol{a}}{dt} & =-\frac{\partial H}{\partial\boldsymbol{x}}\,.
\end{align}
The Hamiltonian (\ref{eq:hmiltonian}) is the same as in Pontryagin's principle
\cite{Pontryagin2018}.

The Euler-Lagrange and the Hamilton form of the equations allows us
easily derive consequences of the possible symmetries in equation
(\ref{eq:diff-general}). For example, if the parameters $\boldsymbol{q}$
do not depend on $t$ then $\partial L/\partial t=0$ and according
to the Noether's theorem \cite{Noether1918} this symmetry leads to
a conservation law 
\begin{equation}
\dot{\boldsymbol{x}}\frac{\partial L}{\partial\dot{\boldsymbol{x}}}-L=
\boldsymbol{a}\boldsymbol{F}(\boldsymbol{x},\boldsymbol{q})=
\mathrm{const}_{t}\,.\label{eq:conservation-fixed-params}
\end{equation}
Therefore
\begin{equation}
\boldsymbol{a}(t)\boldsymbol{F}(\boldsymbol{x}(t),\boldsymbol{q})=
\frac{\partial\mathcal{L}}{\partial\boldsymbol{x}(T)}\boldsymbol{F}(\boldsymbol{x}(T),\boldsymbol{q})\,.
\end{equation}

\subsection{Probability density functions}

Let us assume that initial data $\boldsymbol{x}(0)$ is characterized
by a probability density function $P(\boldsymbol{x}(0),0)$. The solution
of the differential equation (\ref{eq:diff-general}) has a transformed
probability density $P(\boldsymbol{x}(t),t)$. Since the transformation
of the variable $\boldsymbol{x}(t)$ is described by a differential
equation, the probability density $P(\boldsymbol{x},t)$ obeys the
continuity equation \cite{Sonoda2017}
\begin{equation}
\frac{\partial}{\partial t}P(\boldsymbol{x},t)+
\mathrm{Tr}\left\{ \frac{\partial}{\partial\boldsymbol{x}}[P(\boldsymbol{x},t)
\boldsymbol{F}(\boldsymbol{x},\boldsymbol{q}(t))]\right\} =0\,.\label{eq:continuity}
\end{equation}
Computing partial derivatives we obtain the partial differential equation
\begin{equation}
\frac{\partial}{\partial t}P(\boldsymbol{x},t)+
\frac{\partial}{\partial\boldsymbol{x}}P(\boldsymbol{x},t)
\boldsymbol{F}(\boldsymbol{x},\boldsymbol{q}(t))+
\mathrm{Tr}\left\{ \frac{\partial}{\partial\boldsymbol{x}}
\boldsymbol{F}(\boldsymbol{x},\boldsymbol{q}(t))\right\} P(\boldsymbol{x},t)=0\,.\label{eq:diff-prob-dens}
\end{equation}
This equation can be written in the form of the instantaneous change
of variables formula of \cite{Chen2018}:
\begin{equation}
\frac{\partial}{\partial t}\log P(\boldsymbol{x}(t),t)=
-\mathrm{Tr}\left\{ \frac{\partial}{\partial\boldsymbol{x}}
\boldsymbol{F}(\boldsymbol{x},\boldsymbol{q}(t))\right\} \,.
\end{equation}

Now let us consider the probability density $P(\boldsymbol{x},\boldsymbol{a},t)$
of features $\boldsymbol{x}$ together with the gradient $\boldsymbol{a}$
of the loss. From the existence of the Hamiltonian $H(\boldsymbol{x},\boldsymbol{a},t)=
\boldsymbol{a}\boldsymbol{F}(\boldsymbol{x},\boldsymbol{q}(t))$
follows that the Liouville's theorem holds: the total derivative of
the probability density function $P(\boldsymbol{x},\boldsymbol{a},t)$
is zero:
\begin{equation}
\frac{d}{dt}P(\boldsymbol{x},\boldsymbol{a},t)=0
\end{equation}
or, explicitly,
\begin{equation}
\frac{\partial}{\partial t}P(\boldsymbol{x},\boldsymbol{a},t)+
\frac{\partial}{\partial\boldsymbol{x}}P(\boldsymbol{x},\boldsymbol{a},t)
\frac{\partial}{\partial\boldsymbol{a}}[\boldsymbol{a}
\boldsymbol{F}(\boldsymbol{x},\boldsymbol{q}(t))]-
\frac{\partial}{\partial\boldsymbol{x}}[\boldsymbol{a}
\boldsymbol{F}(\boldsymbol{x},\boldsymbol{q}(t))]
\frac{\partial}{\partial\boldsymbol{a}}P(\boldsymbol{x},\boldsymbol{a},t)=0\,.
\end{equation}
The distribution $P(\boldsymbol{x},\boldsymbol{a},t)$ is constant
along any trajectory in phase space $(\boldsymbol{x},\boldsymbol{a})$.

The information entropy associated with the data $\boldsymbol{x}$
is defined as
\begin{equation}
\mathcal{H}(t)=-\int P(\boldsymbol{x},t)\log P(\boldsymbol{x},t)\,d^{N}\boldsymbol{x}\,.
\end{equation}
Since $P(\boldsymbol{x},t)$ obeys the differential equation (\ref{eq:diff-prob-dens}),
for the information entropy we have
\begin{equation}
\frac{d}{dt}\mathcal{H}(t)=
\int\mathrm{Tr}\left\{ \frac{\partial}{\partial\boldsymbol{x}}
\boldsymbol{F}(\boldsymbol{x},\boldsymbol{q}(t))\right\} P(\boldsymbol{x},t)\,d^{N}\boldsymbol{x}\,.
\end{equation}
With increasing $t$ the model should discard the features that are
unnecessary for the prediction of true label $\boldsymbol{y}_{\mathrm{true}}$.
Thus one can argue that the information entropy should decrease with
$t$ , $d\mathcal{H}(t)/dt\leqslant0$. This also follows from the
requirement of the stability of differential equations (\ref{eq:diff-general})
\cite{Haber2018}.

\subsection{Symmetric equations\label{subsec:symmetric}}

Let us consider the case where the function $\boldsymbol{F}(\boldsymbol{x},\boldsymbol{q})$
satisfies the condition
\begin{equation}
\frac{\partial}{\partial\boldsymbol{x}}\boldsymbol{F}(\boldsymbol{x},\boldsymbol{q})=
\frac{\partial}{\partial\boldsymbol{x}^{\intercal}}\boldsymbol{F}(\boldsymbol{x},\boldsymbol{q})^{\intercal}\,,
\end{equation}
that is, the matrix $\partial\boldsymbol{F}/\partial\boldsymbol{x}$
is symmetric. Then the vector field $\boldsymbol{F}$ is conservative
and, therefore, it can be expressed as a gradient of of scalar function:
\begin{equation}
\boldsymbol{F}(\boldsymbol{x},\boldsymbol{q})=
-\frac{\partial}{\partial\boldsymbol{x}^{\intercal}}E(\boldsymbol{x},\boldsymbol{q})\,.
\end{equation}
In such a case the differential equation (\ref{eq:diff-general})
can be written as
\begin{equation}
\frac{d\boldsymbol{x}}{dt}=
-\frac{\partial}{\partial\boldsymbol{x}^{\intercal}}E(\boldsymbol{x},\boldsymbol{q})\,.
\end{equation}
If the parameters $\boldsymbol{q}$ do not depend on $t$, the function
$E(\boldsymbol{x}(t),\boldsymbol{q})$ does not increase with with
$t$. Indeed, we have
\begin{equation}
\frac{d}{dt}E(\boldsymbol{x},\boldsymbol{q})=
\frac{\partial}{\partial\boldsymbol{x}}E(\boldsymbol{x},\boldsymbol{q})
\frac{d\boldsymbol{x}}{dt}=-\frac{d\boldsymbol{x}^{\intercal}}{dt}\frac{d\boldsymbol{x}}{dt}\leq0\,.
\end{equation}
The existence of an energy function that decrease along trajectories
is a property of Cohen-Grossberg model \cite{Cohen1983} and Hopfield's
\cite{Hopfield1984} continuous time recurrent neural networks.

\subsection{Gradient descent updates of the parameters}

If the parameters $\boldsymbol{q}(t)$ are changed by $\Delta\boldsymbol{q}(t)$,
the change of the loss function $\mathcal{L}$ due to the change of
the parameters is
\begin{equation}
\Delta\mathcal{L}=\int_{0}^{T}\frac{\delta\mathcal{L}}{\delta\boldsymbol{q}(t)}\Delta\boldsymbol{q}(t)\,dt\,.
\end{equation}
Let us find the optimal updates of the parameters in the gradient
descent step, leading to a largest change $\Delta\mathcal{L}$ of
the loss. In order to fix a learning rate we will keep the $L_{2}$
norm of $\Delta\boldsymbol{q}(t)$ fixed,
$\int_{0}^{T}\Delta\boldsymbol{q}(t)^{\intercal}\Delta\boldsymbol{q}(t)\,dt=1$.
Then the optimal change $\Delta\boldsymbol{q}(t)$ can be found using
Lagrange multipliers by maximizing the functional
\begin{equation}
\mathcal{I}[\Delta\boldsymbol{q}]=
\int_{0}^{T}\frac{\delta\mathcal{L}}{\delta\boldsymbol{q}(t)}\Delta\boldsymbol{q}(t)\,dt+
\lambda\left(1-\int_{0}^{T}\Delta\boldsymbol{q}(t)^{\intercal}\Delta\boldsymbol{q}(t)\,dt\right)\,.
\end{equation}
This gives the optimal change
\begin{equation}
\Delta\boldsymbol{q}(t)=-\alpha\frac{\delta\mathcal{L}}{\delta\boldsymbol{q}(t)^{\intercal}}\label{eq:change-q}
\end{equation}
leading to the change of the loss function
\begin{equation}
\Delta\mathcal{L}=-\alpha\int_{0}^{T}\frac{\delta\mathcal{L}}{\delta\boldsymbol{q}(t)}
\frac{\delta\mathcal{L}}{\delta\boldsymbol{q}(t)^{\intercal}}\,dt\,.\label{eq:change-1}
\end{equation}
Using equation (\ref{eq:func-deriv-3}) we get the updates of the
parameters in the gradient descent step
\begin{equation}
\boldsymbol{q}^{\prime}(t)=\boldsymbol{q}(t)-
\alpha\frac{\partial}{\partial\boldsymbol{q}(t)^{\intercal}}
\boldsymbol{F}(\boldsymbol{x}(t),\boldsymbol{q}(t))^{\intercal}\boldsymbol{a}(t)^{\intercal}\,.
\end{equation}

When the function $\boldsymbol{q}(t)$ is parameterized by parameters
$\boldsymbol{\theta}$, then the change is
\begin{equation}
\Delta\boldsymbol{q}(t)=\frac{\partial\boldsymbol{q}(t)}{\partial\theta_{i}}\Delta\theta_{i}
\end{equation}
with
\begin{equation}
\Delta\theta_{i}=-\alpha'\frac{\partial\mathcal{L}}{\partial\theta_{i}}=
-\alpha'\int_{0}^{T}\frac{\delta\mathcal{L}}{\delta\boldsymbol{q}(t)}
\frac{\partial\boldsymbol{q}(t)}{\partial\theta_{i}}\,dt\,.
\end{equation}
The change of the loss function becomes
\begin{equation}
\Delta\mathcal{L}=-\alpha'\sum_{i}\left(\int_{0}^{T}\frac{\delta\mathcal{L}}{\delta\boldsymbol{q}(t)}
\frac{\partial\boldsymbol{q}(t)}{\partial\theta_{i}}\,dt\right)^{2}\,.
\end{equation}
In particular, if the parameters $\boldsymbol{q}$ do net depend on
$t$ then
\begin{equation}
\Delta\boldsymbol{q}=-\alpha'\int_{0}^{T}\frac{\delta\mathcal{L}}{\delta\boldsymbol{q}(t)^{\intercal}}\,dt
\end{equation}
and the change of the loss function is
\begin{equation}
\Delta\mathcal{L}=-\alpha'\int_{0}^{T}\frac{\delta\mathcal{L}}{\delta\boldsymbol{q}(t')}\,dt'
\int_{0}^{T}\frac{\delta\mathcal{L}}{\delta\boldsymbol{q}(t)^{\intercal}}\,dt\,.\label{eq:change-2}
\end{equation}
One can see that in the equation (\ref{eq:change-1}) the integral
is taken of the non-negative quantity. Each infinitesimal segment
$\delta t$ has a positive contribution to the change of the loss.
On the other hand, in equation (\ref{eq:change-2}) the integration
is performed of the quantities that can have various signs and different
segments of $t$ can partially cancel each other, resulting in smaller
change of the loss. The parameters $\boldsymbol{q}(t)$ that depend
on $t$ can lead to faster training of the model.

\section{Ordinary differential equation as a model of feedforward neural networks\label{sec:feedforward-ode}}

\subsection{Heuristic derivation of differential equation}

In this section we present a non-strict derivation of a differential
equation that captures essential properties of feedforward neural
networks. Let us consider a feedforward neural network consisting
of a large number $L$ of layers. The $l$-th layer ($l\in\{1,2,\ldots,L\}$)
applies a nonlinear transform on its input $\boldsymbol{x}^{(l)}$
to produce its output $\boldsymbol{x}^{(l+1)}$, where the nonlinear
transform is an affine transform
\begin{equation}
\boldsymbol{u}^{(l+1)}=\boldsymbol{W}^{(l)}\boldsymbol{x}^{(l)}+\boldsymbol{B}^{(l)}\label{eq:affine}
\end{equation}
followed by a non-linear activation function $h(\boldsymbol{x})$:
\begin{equation}
\boldsymbol{x}^{(l+1)}=h(\boldsymbol{u}^{(l+1)})\,.\label{eq:nonlin}
\end{equation}
Here in the equation (\ref{eq:affine}) $\boldsymbol{W}^{(l)}$ is
a matrix of weights and $\boldsymbol{B}^{(l)}$ is a bias vector.
For simplicity at first we assume that the vector $\boldsymbol{x}^{(l+1)}$
has the same number of dimensions as $\boldsymbol{x}^{(l)}$; the
situation when the dimensions are different will be discussed below
(subsection \ref{subsec:diff-dimen}). In order to obtain a differential
equation we require that the vector $\boldsymbol{x}^{(l+1)}$ differ
only slightly from $\boldsymbol{x}^{(l)}$. Small difference between
the input and the output of a layer can occur when the matrix $\boldsymbol{W}^{(l)}$
has the form
\begin{equation}
\boldsymbol{W}^{(l)}=\boldsymbol{I}+\Delta t\boldsymbol{w}^{(l)}\,,\label{eq:weight-structure}
\end{equation}
where $\boldsymbol{I}$ is an identity matrix and $\Delta t\ll1$
is a small parameter. In addition, the bias also should be proportional
to the small parameter,
\begin{equation}
\boldsymbol{B}^{(l)}=\Delta t\boldsymbol{b}^{(l)}\,.
\end{equation}
In such a case the affine transform reads
\begin{equation}
\boldsymbol{u}^{(l+1)}=\boldsymbol{x}^{(l)}+\Delta t(\boldsymbol{w}^{(l)}\boldsymbol{x}^{(l)}+
\boldsymbol{b}^{(l)})\,.\label{eq:affine-small}
\end{equation}
Thus the affine transform is a sum of two terms, the first term being
an unchanged input and the second term being proportional to the small
parameter $\Delta t$. The first term resembles a shortcut connection
in highway networks \cite{Srivastava2015,Srivastava2015a} and residual
networks \cite{He2016}.

Furthermore, to get small difference between the input and the output
of a layer, the non-linear function $h(\boldsymbol{u})$ should be
almost linear. We can obtain such a nonlinear function by adding a
small cubic term to the linear function:
\begin{equation}
h(u)=u-\gamma\Delta tu^{3}\,,\label{eq:nonlin-small}
\end{equation}
where $\gamma>0$ is a parameter of nonlinearity. One can interpret
this non-linear function as a generalization of a hyperbolic tangent
activation, because the first two terms in the Taylor series of $\tanh x$
are $\tanh x=x-x^{3}/3$. Note, that Taylor series of arbitrary function
$h(x)$ should have also a quadratic term $x^{2}$, however the quadratic
term can be eliminated by the shift of $x$. If one considers non-analytic
functions, equation (\ref{eq:nonlin}) can be generalized as
\begin{equation}
h(u)=u-\gamma\Delta t\sign(u)|u|^{\mu}\,,\label{eq:nonlin-general}
\end{equation}
where $\mu>1$ is the exponent of nonlinearity.

Combining the equations (\ref{eq:nonlin}), (\ref{eq:affine-small}),
(\ref{eq:nonlin-small}) and keeping only the terms up to the first
order in $\Delta t$ we obtain
\begin{equation}
x_{i}^{(l+1)}=x_{i}^{(l)}+\Delta t\left(\sum_{j}w_{i,j}^{(l)}x_{j}^{(l)}+b_{i}^{(l)}
-\gamma x_{i}^{(l)3}\right)\,.\label{eq:transform-small}
\end{equation}
Instead of the layer number $l$ we will use $t=l\Delta t$ and write
$\boldsymbol{x}(t=l\Delta t)$ instead of $\boldsymbol{x}^{(l)}$
, rewriting the equation (\ref{eq:transform-small}) as
\begin{equation}
\frac{1}{\Delta t}[x_{i}(t+\Delta t)-x_{i}(t)]=\sum_{j}w_{i,j}(t)x_{j}(t)+b_{i}(t)-\gamma x_{i}(t)^{3}\,.
\end{equation}
Interpreting $t$ as a continuous variable and taking the limit $\Delta t\rightarrow0$
we obtain a differential equation
\begin{equation}
\frac{d}{dt}x_{i}(t)=\sum_{j}w_{i,j}(t)x_{j}(t)+b_{i}(t)-\gamma x_{i}(t)^{3}\,.\label{eq:main}
\end{equation}
Note, that the equation (\ref{eq:main}) is non-linear, the non-linear
term $\gamma x^{3}$ plays the role of a non-linear activation function.
Using the non-linear activation function (\ref{eq:nonlin-general})
the differential equation becomes
\begin{equation}
\frac{d}{dt}x_{i}(t)=\sum_{j}w_{i,j}(t)x_{j}(t)+b_{i}(t)-
\gamma\sign[x_{i}(t)]|x_{i}(t)|^{\mu}\,.\label{eq:forward-general}
\end{equation}
Since the non-linear differential equation (\ref{eq:main}) has been
obtained starting from a discrete neural network, we expect that the
proposed equation can reflect some properties of deep neural networks.

Equation (\ref{eq:main}) can be further generalized to the case of
continuous indices, yielding an integro-differential equation
\begin{equation}
\frac{\partial}{\partial t}x(u,t)=\int w(u,v,t)x(v,t)\,dv+b(u,t)-
\gamma x(u,t)^{3}\,.\label{eq:forward-continuous}
\end{equation}
Such double continuum limit of neural networks has been proposed in
\cite{Sonoda2017}. Taking a singular kernel $w(u,v,t)$ in equation
(\ref{eq:forward-continuous}) one can get a partial differential
equation instead of an integral one \cite{Ruthotto2018,Long2017,Long2018}.

\subsection{Different number of dimensions in input and output\label{subsec:diff-dimen}}

When the number of dimensions $N^{(l+1)}$ in the output $\boldsymbol{x}^{(l+1)}$
differs form the number of dimensions $N^{(l)}$ in the input $\boldsymbol{x}^{(l)}$,
the structure of the weights (\ref{eq:weight-structure}) is not possible
and the transition to the differential equation is not straightforward.
However, when the number of dimensions is increased, $N^{(l+1)}>N^{(l)}$
at some $l=L_{0}$, we can interpret this increase as the presence
of $N^{(l+1)}-N^{(l)}$ units with the activation equal to zero in
the $l$-th layer. Then we can require the structure of the weights
$\boldsymbol{W}^{(l)}$ that can be described by an equation similar
to (\ref{eq:weight-structure}), where the identity matrix $\boldsymbol{I}$
is replaced by a rectangular matrix with ones on the main diagonal
and zeros elsewhere. Equation (\ref{eq:transform-small}) then becomes
\begin{equation}
x_{i}^{(l+1)}=\begin{cases}
x_{i}^{(l)}+\Delta t\sum_{j}w_{i,j}^{(l)}x_{j}^{(l)}+\Delta tb_{i}^{(l)}-
\Delta t\gamma x_{i}^{(l)3}\,, & i\leqslant N^{(l)}\,,\\
\Delta t\sum_{j}w_{i,j}^{(l)}x_{j}^{(l)}+\Delta tb_{i}^{(l)}\,, & i>N^{(l)}\,.
\end{cases}
\end{equation}
This equation keeps the $L_{2}$ norm of $\boldsymbol{x}^{(l+1)}$
equal to the norm of $\boldsymbol{x}^{(l)}$ when $\Delta t=0$. In
the limit $\Delta t\rightarrow0$ we get the differential equation
(\ref{eq:main}) with an additional boundary condition $x_{i}(t=T_{0})=0$
if $i>N^{(L_{0})}$.

When the number of dimensions in the output is decreased, a rectangular
matrix $\boldsymbol{W}^{(l)}$ that keeps the $L_{2}$ norm of a vector
$\boldsymbol{W}^{(l)}\boldsymbol{x}^{(l)}$ equal to the norm of $\boldsymbol{x}^{(l)}$
does not exist. Thus decreasing the number of units necessarily leads
to the change of the norm of the signal.

\subsection{Backward propagation for the proposed model\label{subsec:backward}}

The general equations obtained in the previous section \ref{sec:odes}
can be applied to the proposed differential equation (\ref{eq:main}),
which corresponds to the function
\begin{equation}
F_{i}(\boldsymbol{x}(t),\boldsymbol{w}(t),\boldsymbol{b}(t))=
\sum_{j}w_{i,j}(t)x_{j}(t)+b_{i}(t)-\gamma x_{i}(t)^{3}\,.
\end{equation}
Thus,
\begin{equation}
\frac{\partial}{\partial x_{j}(t)}F_{i}(\boldsymbol{x}(t),\boldsymbol{w}(t),\boldsymbol{b}(t))=
w_{i,j}(t)-3\gamma\delta_{i,j}x_{j}(t)^{2}
\end{equation}
and the differential equation for the backward propagation becomes
\begin{equation}
-\frac{d}{dt}a_{i}(t)=\sum_{j}a_{j}(t)w_{j,i}(t)-3\gamma x_{i}(t)^{2}a_{i}(t)\,.\label{eq:main-backward}
\end{equation}
The last term on the right hand side of equation (\ref{eq:main-backward})
represents a decay of the gradient with the decay rate $3\gamma x_{i}^{2}$
that increases with increasing signal $x_{i}$. In a similar way,
starting with the forward propagation equation (\ref{eq:forward-general}),
we get the differential equation
\begin{equation}
-\frac{d}{dt}a_{i}(t)=\sum_{j}a_{j}(t)w_{j,i}(t)-\mu\gamma|x_{i}(t)|^{\mu-1}a_{i}(t)\,.
\end{equation}
In the case of continuous indices this backward propagation equation
has the form 
\begin{equation}
-\frac{\partial}{\partial t}a(u,t)=\int a(v,t)w(v,u,t)\,dv-3\gamma x(u,t)^{2}a(u,t)\,.
\end{equation}

According to the equation (\ref{eq:func-deriv-3}), the gradients
(functional derivatives) of the loss function $\mathcal{L}$ with
respect to the parameters $\boldsymbol{w}(t)$ and $\boldsymbol{b}(t)$
are
\begin{align}
\frac{\delta\mathcal{L}}{\delta\boldsymbol{w}(t)} & =\boldsymbol{x}(t)\boldsymbol{a}(t)\,,\\
\frac{\delta\mathcal{L}}{\delta\boldsymbol{b}(t)} & =\boldsymbol{a}(t)\,.
\end{align}
These gradients lead to the updates of the parameters in the gradient
descent step
\begin{align}
\boldsymbol{w}^{\prime}(t) & =
\boldsymbol{w}(t)-\alpha\boldsymbol{a}(t)^{\intercal}\boldsymbol{x}(t)^{\intercal}\,,\label{eq:update-w}\\
\boldsymbol{b}^{\prime}(t) & =\boldsymbol{b}(t)-\alpha\boldsymbol{a}(t)^{\intercal}\,.\label{eq:update-b}
\end{align}
Here the function $\boldsymbol{a}(t)$ should be obtained by solving
equation~(\ref{eq:main-backward}) first. However, from equations
(\ref{eq:main}) and (\ref{eq:main-backward}) follows that we can
obtain the updates to the parameters $\delta w_{i,j}(t)=a_{i}(t)x_{j}(t)$
and $\delta b_{i}(t)=a_{i}(t)$ by directly solving the following
differential equations:
\begin{align}
\frac{d}{dt}\delta w_{i,j}(t) & =\sum_{k}[w_{j,k}(t)\delta w_{i,k}(t)-
w_{k,i}(t)\delta w_{k,j}(t)]+b_{j}(t)\delta b_{i}(t)+
\gamma[3x_{i}(t)^{2}-x_{j}(t)^{2}]\delta w_{i,j}(t)\,,\\
\frac{d}{dt}\delta b_{i}(t) & =-\sum_{j}w_{j,i}(t)\delta b_{j}(t)+
3\gamma x_{i}(t)^{2}\delta b_{i}(t)\,.
\end{align}

We can consider the parameter of nonlinearity $\gamma$ as an additional
parameter to be learned instead of having a fixed value. The learnable
parameter of nonlinearity is similar to the idea of Parametric ReLUs
\cite{He2015}. From the equation (\ref{eq:grdient-fixed-param})
we can obtain the gradient of the loss with respect to the parameter
$\gamma$:
\begin{equation}
\frac{\partial\mathcal{L}}{\partial\gamma}=-\sum_{i}\int_{0}^{T}a_{i}(t)x_{i}(t)^{3}\,dt\,.
\end{equation}

\subsection{Convolutional neural networks\label{subsec:convolutional}}

Equation (\ref{eq:main}) can be easily adapted to model convolutional
neural networks. For convolutional neural networks the affine transformation
(\ref{eq:affine}) reads
\begin{equation}
u_{i}^{(l+1)}=\sum_{j}K_{j}^{(l)}x_{i+j}^{(l)}+B^{(l)}\,.
\end{equation}
Here for simplicity we write the equation as one-dimensional cross-correlation
with only one feature channel. The generalization to more dimensions
and more channels is straightforward. In addition, we assume that
input $\boldsymbol{x}^{(l)}$ is zero padded so that the size of the
output is the same as the size of the input. Requiring that
\begin{equation}
K_{i}^{(l)}=\delta_{i,0}+\Delta t\kappa_{i}^{(l)}
\end{equation}
and
\begin{equation}
B^{(l)}=\Delta tb^{(l)}
\end{equation}
in the limit $\Delta t\rightarrow0$ we obtain the differential equation
\begin{equation}
\frac{d}{dt}x_{i}(t)=\sum_{j}\kappa_{j}(t)x_{i+j}(t)+b(t)-
\gamma x_{i}(t)^{3}\,.\label{eq:conv-forward}
\end{equation}
When there are more than one channel we assume that the number of
channels in each layer remains the same. In this case the forward
propagation is described by differential equations that are similar
to the equations (\ref{eq:conv-forward}), only with summation over
channels added.

Proceeding similarly as in subsection \ref{subsec:backward}, from
equation (\ref{eq:conv-forward}) we get the differential equation
\begin{equation}
-\frac{d}{dt}a_{i}(t)=\sum_{j}a_{i-j}(t)\kappa_{j}(t)-
3\gamma x_{i}(t)^{2}a_{i}(t)\label{eq:conv-backward}
\end{equation}
describing the backward propagation for the convolutional networks.
When there are more than one channel, the backward propagation is
described by differential equations that are similar to the equations
(\ref{eq:conv-backward}) with added summation over channels. Finally,
the gradients of the loss function with respect to the parameters
are
\begin{align}
\frac{\delta\mathcal{L}}{\delta\kappa_{i}(t)} & =\sum_{j}a_{j}(t)x_{j+i}(t)\,,\label{eq:grad-kappa}\\
\frac{\delta\mathcal{L}}{\delta b(t)} & =\sum_{j}a_{j}(t)\,,\label{eq:grad-b}
\end{align}
and lead to the updates of the parameters in the gradient descent
step
\begin{align}
\kappa_{i}^{\prime}(t) & =\kappa_{i}(t)-\alpha\sum_{j}a_{j}(t)x_{j+i}(t)\,,\\
b^{\prime}(t) & =b(t)-\alpha\sum_{j}a_{j}(t)\,.
\end{align}
Here the functions $a_{j}(t)$ are the solutions of equation (\ref{eq:conv-backward}).

\subsection{Regularization\label{subsec:regularization}}

Deep neural networks with a large number of learnable parameters are
prone to overfitting when trained on a relatively small training set.
The learned weights become adjusted only for the training set and
the network lacks the generalization ability to the unseen data. One
of the causes of the the overfitting problem is the co-adaptation
of the neurons that results in the neural network dependent on the
joint response instead of each neuron learning a useful feature representation
\cite{Hinton2012}. To prevent overfitting during the training of
the network a number of regularization methods have been introduced.
These methods include data augmentation \cite{Simard2003}, $L_{1}$
and $L_{2}$ regularization \cite{Montavon2012}, early stopping \cite{Morgan1990},
and random dropout \cite{Srivastava2014}. Description of the forward
propagation by a differential equation provides yet another method
of regularization: one can prefer the weights that vary smoothly with
$t$. The variation of the parameters can be minimized by using the
regularization functional

\begin{equation}
R(\boldsymbol{w},\boldsymbol{b})=\frac{1}{2}\int_{0}^{T}\left(\left\Vert \frac{d}{dt}\boldsymbol{w}\right\Vert _{\mathrm{F}}^{2}
+\left\Vert \frac{d}{dt}\boldsymbol{b}\right\Vert _{2}^{2}\right)\,dt\,,
\end{equation}
where $\left\Vert \cdot\right\Vert _{\mathrm{F}}$ represents the
Frobenius norm. Discretization with the step $\Delta t$ yields the
regularization function
\begin{equation}
R_{\Delta t}=\frac{1}{2\Delta t}\sum_{l=2}^{L}\left(\left\Vert \boldsymbol{w}^{(l)}-
\boldsymbol{w}^{(l-1)}\right\Vert _{\mathrm{F}}^{2}+
\left\Vert \boldsymbol{b}^{(l)}-\boldsymbol{b}^{(l-1)}\right\Vert _{\mathrm{2}}^{2}\right)\,.
\end{equation}
Such a regularization has been proposed in \cite{Haber2018,Chang2017,Ruthotto2018}.
This method of regularization has been justified by the stability
requirements in \cite{Haber2018}.

\section{Properties of the proposed differential equation\label{sec:properties}}

In this section we will examine the differential equation (\ref{eq:main}),
proposed in the previous section. This equation has several scaling
properties. We can eliminate the parameter $\gamma$ by rescaling
the variable $t$ to a scaled variable $t_{\mathrm{s}}=\gamma t$
and introducing scaled weights and biases,
$\boldsymbol{w}_{\mathrm{s}}=\gamma^{-1}\boldsymbol{w}$,
$\boldsymbol{b}_{\mathrm{s}}=\gamma^{-1}\boldsymbol{b}$. Therefore,
without losing generality one can set $\gamma=1$. In addition, rescaled
activations $\boldsymbol{x}_{\mathrm{s}}=c\boldsymbol{x}$, where
$c$ is some constant, obey the same equation (\ref{eq:main}) only
with scaled weights and biases $\boldsymbol{w}_{\mathrm{s}}=c^{2}\boldsymbol{w}$,
$\boldsymbol{b}_{\mathrm{s}}=c^{3}\boldsymbol{b}$ and scaled variable
$t_{\mathrm{s}}=c^{-2}t$. It follows that the characteristic scale
in the dependence of the solutions of the equation (\ref{eq:main})
on $t$ is influenced by the norm of the vector $\boldsymbol{x}$.

\subsection{Some solutions of the proposed equation}

It is instructive to solve the differential equation (\ref{eq:main})
for a some particular values of the parameters $\boldsymbol{w}$,
$\boldsymbol{b}$. In all the cases we will take symmetric matrix
$\boldsymbol{w}$, $\boldsymbol{w}^{\intercal}=\boldsymbol{w}$ and
consider parameters that do not depend on $t$. Then, according to
the results of section~\ref{subsec:symmetric}, the quantity
\begin{equation}
E(\boldsymbol{x})=-\frac{1}{2}\boldsymbol{x}^{\intercal}\boldsymbol{w}\boldsymbol{x}-
\boldsymbol{b}^{\intercal}\boldsymbol{x}+\frac{1}{4}\gamma\boldsymbol{x}^{4}\label{eq:main-energy}
\end{equation}
does not increase with with $t$. The solution of the differential
equation approach minima of $E(\boldsymbol{x})$ when $t\rightarrow\infty$.

When the weights matrix $\boldsymbol{w}$ is diagonal and does not
depend on $t$, biases are zero $\boldsymbol{b}(t)=0$, the forward
propagation is described by the equation
\begin{equation}
\frac{d}{dt}x_{i}(t)=w_{i,i}x_{i}(t)-\gamma x_{i}(t)^{3}\,.
\end{equation}
The solution obeying the initial condition $x_{i}(t=0)=x_{i}(0)$
is
\begin{equation}
x_{i}(t)=\frac{x_{i}(0)e^{w_{i,i}t}}{\sqrt{1+
\frac{\gamma}{w_{i,i}}x_{i}(0)^{2}(e^{2w_{i,i}t}-1)}}\label{eq:a-wc-b0}
\end{equation}
If $w_{i,i}<0$, at large values of $t$, $t\gg|w_{i,i}|^{-1}$, the
solution (\ref{eq:a-wc-b0}) exponentially approaches zero. If $w_{i,i}>0$,
at large values of $t$ depending on the sign of the initial value
$x_{i}(0)$ the solution approaches one of the two fixed values
\begin{equation}
x_{i}(t)\approx\sign x_{i}(0)\sqrt{\frac{w_{i,i}}{\gamma}}\,.
\end{equation}
In the particular case of $w_{i,i}=0$ the solution (\ref{eq:a-wc-b0})
is
\begin{equation}
x_{i}(t)=\frac{x_{i}(0)}{\sqrt{1+2\gamma x_{i}(0)^{2}t}}\,.\label{eq:a-w0-b0}
\end{equation}
At large values of $t$, $t\gg\gamma^{-1}x_{i}(0)^{-1}$, the solution
(\ref{eq:a-w0-b0}) approaches
\begin{equation}
x_{i}(t)\approx\sign x_{i}(0)\frac{1}{\sqrt{2\gamma t}}\,.\label{eq:a-asym}
\end{equation}
The activation decays to zero but as a power-law, $t^{-1/2}$, not
exponentially. We can interpret the model with the parameters equal
to zero as a situation when the model does not have any information
about the possible inputs. Inserting the solution (\ref{eq:a-wc-b0})
into the backward propagation equation (\ref{eq:main-backward}) we
obtain the gradients
\begin{equation}
a_{i}(t)=a_{i}(T)e^{w_{i,i}(T-t)}\left(\frac{1+\frac{\gamma}{w_{i,i}}x_{i}(0)^{2}(e^{2w_{i,i}t}-1)}{1+
\frac{\gamma}{w_{i,i}}x_{i}(0)^{2}(e^{2w_{i,i}T}-1)}\right)^{\frac{3}{2}}\,.\label{eq:g-wc-b0}
\end{equation}
This solution can be obtained without solving the adjoint equation
by utilizing the conservation law (\ref{eq:conservation-fixed-params}).
When $w_{i,i}<0$, the gradient at large $t$ decays exponentially
with decreasing $t$ as $a_{i}(t)=a_{i}(T)e^{-|w_{i,i}|(T-t)}$. When
$w_{i,i}>0$, the gradient also decays exponentially, as
$a_{i}(t)\approx a_{i}(T)e^{-2w_{i,i}(T-t)}$.
When $w_{i,i}=0$, equation (\ref{eq:g-wc-b0}) becomes
\begin{equation}
a_{i}(t)=a_{i}(T)\left(\frac{1+2\gamma z_{i}(0)^{2}t}{1+
2\gamma z_{i}(0)^{2}T}\right)^{\frac{3}{2}}\,.\label{eq:g-w0-b0}
\end{equation}
This equation shows that the gradients decay as a power-law function
of $t$ when $t$ decreases toward zero.

When weights are zero, $\boldsymbol{w}(t)=0$ and biases $\boldsymbol{b}$
do not depend on $t$, the solution of the equation for the forward
propagation
\begin{equation}
\frac{d}{dt}x_{i}(t)=b_{i}-\gamma x_{i}(t)^{3}
\end{equation}
with increasing $t$ approaches a stationary value
\begin{equation}
x_{i}(\infty)=\gamma^{-\frac{1}{3}}b_{i}^{\frac{1}{3}}\,.\label{eq:a-stationary}
\end{equation}
This means that the backward propagation equation (\ref{eq:main-backward})
at large $t$ becomes
\begin{equation}
\frac{d}{dt}a_{i}(t)\approx3\gamma^{\frac{1}{3}}b_{i}^{\frac{2}{3}}a_{i}(t)
\end{equation}
which has a solution
\begin{equation}
a_{i}(t)\approx a_{i}(T)e^{-3\gamma^{\frac{1}{3}}b_{i}^{\frac{2}{3}}(T-t)}
\end{equation}
We see that in this case the gradient decays exponentially with decreasing
$t$.

\subsection{Effect of weights matrix $\boldsymbol{w}$\label{subsec:weights}}

The weights matrix $\boldsymbol{w}(t)$ in the first term of the equation
(\ref{eq:main}) can be separated into symmetric and antisymmetric
(or skew-symmetric) parts:
\begin{equation}
\boldsymbol{w}(t)=\boldsymbol{w}_{\mathrm{sym}}(t)+
\boldsymbol{w}_{\mathrm{anti}}(t)\,,
\end{equation}
where
\begin{align}
\boldsymbol{w}_{\mathrm{sym}}(t) & =\frac{1}{2}(\boldsymbol{w}(t)+
\boldsymbol{w}(t)^{\intercal})\,,\\
\boldsymbol{w}_{\mathrm{anti}}(t) & =\frac{1}{2}(\boldsymbol{w}(t)-
\boldsymbol{w}(t)^{\intercal})\,.
\end{align}
The matrix transposition operation leaves the symmetric matrix unchanged,
$\boldsymbol{w}_{\mathrm{sym}}^{\intercal}=\boldsymbol{w}_{\mathrm{sym}}$,
the antisymmetrix matrix changes the sign,
$\boldsymbol{w}_{\mathrm{anti}}^{\intercal}=-\boldsymbol{w}_{\mathrm{anti}}$.
In terms of the matrix elements these properties read
$(w_{\mathrm{sym}})_{i,j}=(w_{\mathrm{sym}})_{j,i}$
and $(w_{\mathrm{anti}})_{i,j}=-(w_{\mathrm{anti}})_{j,i}$. In convolutional
neural networks the antisymmetric property of the weights matrix corresponds
to the kernel that changes sign when the indices are reversed,
\begin{equation}
\kappa_{-i}=-\kappa_{i}\,.
\end{equation}
The symmetric part of the weights matrix $\boldsymbol{w}$ can be
diagonalized by a properly chosen orthogonal matrix, that is by a
rotation in a space of activation vectors $\boldsymbol{x}$. If $\boldsymbol{x}(t)$
is an eigenvector of the symmetric matrix $\boldsymbol{w}_{\mathrm{sym}}(t)$
then it is multiplied by the corresponding eigenvalue in equation
(\ref{eq:main}).

The antisymmetric part of the weights matrix causes a rotation in
the $N$-dimensional space of activation vectors around the point
$x_{1}=x_{2}=\cdots=x_{N}=0$; the rotation plane determined by the
coefficients $(w_{\mathrm{anti}})_{i,j}$. This can easily be seen
in the case of two-dimensional activation vectors. Assuming that the
antisymmetric weights matrix $\boldsymbol{w}$ does not depend on
$t$, the bias $\boldsymbol{b}=0$ and the activations are small,
$\gamma x_{i}^{2}\ll1$, so that the term $\gamma x_{i}^{3}$ can
be neglected, the forward propagation equations become
\begin{align}
\frac{d}{dt}x_{1}(t) & =w_{1,2}x_{2}(t)\,,\\
\frac{d}{dt}x_{2}(t) & =-w_{1,2}x_{1}(t)\,.
\end{align}
The solution of the equations is
\begin{align}
x_{1}(t) & =x_{1}(0)\cos(w_{1,2}t)+x_{2}(0)\sin(w_{1,2}t)\,,\\
x_{2}(t) & =x_{2}(0)\cos(w_{1,2}t)-x_{1}(0)\sin(w_{1,2}t)\,.
\end{align}
From this solution we see that $w$ causes rotation of the vector
$\boldsymbol{x}$. In order to obtain stable forward propagation it
was proposed to use only antisymmetric weight matrices in \cite{Haber2018}.

\subsection{Physical interpretation of the proposed equation \label{subsec:physical-interpretation}}

Let us consider the dependence of $L_{2}$ norm of the activation
vector $\boldsymbol{x}$ on the variable $t$. Introducing the quantity
\begin{equation}
E(t)=\frac{1}{2}\boldsymbol{x}(t)^{2}
\end{equation}
that is proportional to the square of the $L_{2}$ norm and using
the equation (\ref{eq:main}) we get the derivative
\begin{align}
\frac{\partial}{\partial t}E(t) & =\boldsymbol{x}(t)^{\intercal}\boldsymbol{w}_{\mathrm{sym}}(t)\boldsymbol{x}(t)+
\boldsymbol{b}(t)^{\intercal}\boldsymbol{x}(t)-\gamma\sum_{i}x_{i}(t)^{4}\label{eq:energy-deriv}\\
 & =P_{w}(t)+P_{b}(t)-P_{\gamma}(t)
\end{align}
where
\begin{align}
P_{w}(t) & =\boldsymbol{x}(t)^{\intercal}\boldsymbol{w}_{\mathrm{sym}}(t)\boldsymbol{x}(t)\label{eq:power-w}\\
P_{\mathrm{b}}(t) & =\boldsymbol{b}(t)^{\intercal}\boldsymbol{x}(t)\,,\\
P_{\gamma}(t) & =\gamma\sum_{i}x_{i}(t)^{4}\,.\label{eq:power-friction}
\end{align}
Equations (\ref{eq:energy-deriv})--(\ref{eq:power-friction}) suggest
the following physical interpretation of the equation (\ref{eq:main}):
we can interpret the vector $\boldsymbol{x}(t)$ as a ``velocity''
in $N$-dimensional space and the variable $t$ as ``time''. The three
terms on the right hand side of the equation (\ref{eq:main}) represent
``forces'' that change the velocity. The first velocity-dependent
term containing the antisymmetric matrix $\boldsymbol{w}_{\mathrm{anti}}(t)$
can be interpreted as an action of a ``magnetic field'' in $N$-dimensional
space, the strength of this magnetic field given by $\boldsymbol{w}_{\mathrm{anti}}(t)$.
Continuing with the electric analogy, the bias vector $\boldsymbol{b}(t)$
can be interpreted as an ``electric field''. The velocity-dependent
term containing the symmetric matrix $\boldsymbol{w}_{\mathrm{sym}}(t)$
represents a linear friction or gain. Finally, the last term $\gamma x_{i}(t)^{3}$
corresponds to a ``friction force'' that is proportional to the velocity
cubed. The quantity $E(t)$ corresponds to the ``kinetic energy'',
whereas the equation (\ref{eq:energy-deriv}) represents the energy
balance: the energy is lost due to non-linear friction with the power
of friction losses given by $P_{\gamma}(t)$ and can be increased
by the action of a ``force'' $\boldsymbol{b}(t)+
\boldsymbol{w}_{\mathrm{sym}}(t)\boldsymbol{x}(t)$
producing the added power $P_{w}(t)+P_{b}(t)$. The ``magnetic field''
$\boldsymbol{w}_{\mathrm{anti}}(t)$ does not change the energy. Note,
that the sign of $\boldsymbol{b}(t)^{\intercal}\boldsymbol{x}(t)$
depends on the angle between the vectors $\boldsymbol{b}(t)$ and
$\boldsymbol{x}(t)$ and can be positive as well as negative.

\subsection{Initialization of weights\label{subsec:initialization}}

For efficient training of deep neural networks a proper initialization
of weights becomes important, since a bad initialization can impede
the learning of a highly non-linear system. The difficulty arises
because in the deep networks the variance of the backpropagated gradient
can vanish or explode due to the multiplicative effects through layers
\cite{Glorot2010}. A proper initialization method should avoid this
exponential reduction or growth of the magnitudes of input signals,
such methods have been proposed in \cite{Glorot2010} and \cite{He2015}.
In our model described by the differential equation (\ref{eq:main})
such an initialization is not necessary and one can set initially
$\boldsymbol{w}(t)=0$. Indeed, as equations (\ref{eq:a-w0-b0}),
(\ref{eq:g-w0-b0}) show, setting $\boldsymbol{w}(t)=0$ leads to
a power-law decay of signals and gradients due to the non-linear activations,
which is much slower than an exponential decay. The initial values
of the weights in the neural networks are usually chosen randomly
(as has been done, for example, in \cite{Krizhevsky2012}) to break
the unwanted initial symmetry. As it is evident from the equations
(\ref{eq:a-w0-b0}), (\ref{eq:g-w0-b0}) and the update rule (\ref{eq:update-w}),
even the initial setting $\boldsymbol{w}(t)=0$ leads to different
updates for different weights thus it is not necessary to start from
unequal weights $\boldsymbol{w}(t)$.

\section{Examples\label{sec:examples}}

The numerical approximation to the solutions of the differential equation
for the forward propagation (\ref{eq:main}) can be obtained using
various numerical methods \cite{Butcher2003}. The simplest is the
forward Euler method, although more complex methods are also possible.
Discretizing the variable $t$ with the step $\Delta t$ and using
the Euler method we obtain the numerical solution scheme given by
the equation (\ref{eq:transform-small}),
\begin{equation}
x_{i}^{(l+1)}=x_{i}^{(l)}+\sum_{j}\tilde{w}_{i,j}^{(l)}x_{j}^{(l)}+
\tilde{b}_{i}^{(l)}-\tilde{\gamma}x_{i}^{(l)3}\,,\label{eq:numer-forward}
\end{equation}
where $x_{i}^{(l)}\equiv x_{i}(l\Delta t)$ and 
\begin{equation}
\tilde{w}_{i,j}^{(l)}=\Delta tw_{i,j}(t_{l})\,,\qquad\tilde{b}_{i}^{(l)}=
\Delta tb_{i}(t_{l})\,,\qquad\tilde{\gamma}=\Delta t\gamma
\end{equation}
are the scaled parameters. Explicit linear multi-step method with
the number of steps equal to two gives the structure \cite{Lu2017,Butcher2003}
\begin{equation}
x_{i}^{(l+1)}=(1-k_{l})x_{i}^{(l)}+k_{l}x_{i}^{(l-1)}+
\sum_{j}\tilde{w}_{i,j}^{(l)}x_{j}^{(l)}+\tilde{b}_{i}^{(l)}-\tilde{\gamma}x_{i}^{(l)3}
\end{equation}
where $k_{l}$ are the trainable parameters for each layer $l$.

\begin{figure}
\centering{}\includegraphics[width=0.3\textwidth]{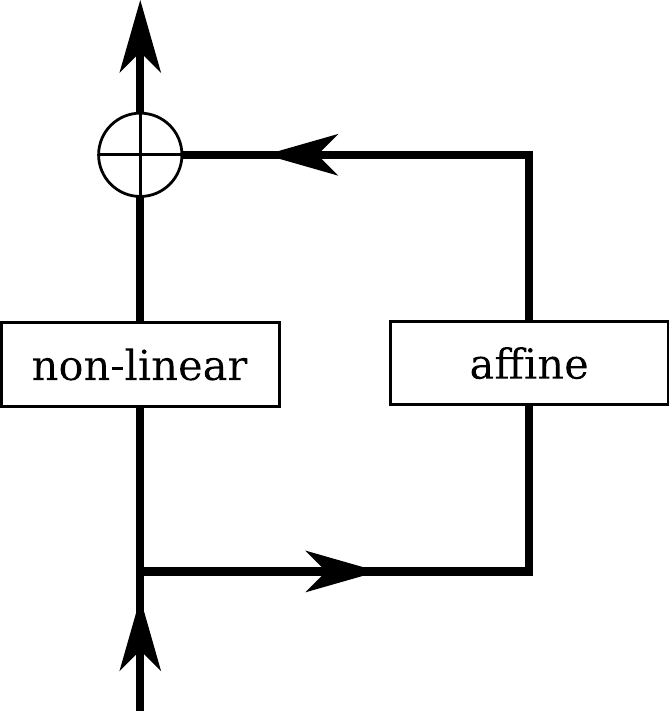}
\caption{Layer structure for the numerical solution of the differential equation
(\ref{eq:main}) using the forward Euler method.}
\label{fig:numer-structure}
\end{figure}

In this section we will investigate the simplest forward Euler method,
given by the equation (\ref{eq:numer-forward}). The numerical solution
scheme for the backward propagation equation (\ref{eq:main-backward})
is
\begin{equation}
a_{i}^{(l)}=a_{i}^{(l+1)}+\sum_{j}a_{j}^{(l+1)}\tilde{w}_{j,i}^{(l)}-
3\tilde{\gamma}x_{i}^{(l)2}a_{i}^{(l+1)}\,.
\end{equation}
The structure of the equation (\ref{eq:numer-forward}) can be graphically
represented by figure \ref{fig:numer-structure}. As equation (\ref{eq:numer-forward})
shows, the output is a sum of a non-linear transformation of the input
$x_{i}^{(l)}-\tilde{\gamma}x_{i}^{(l)3}$ and an affine transformation
$\sum_{j}\tilde{w}_{i,j}^{(l)}x_{j}^{(l)}+\tilde{b}_{i}^{(l)}$. One
can look at the figure \ref{fig:numer-structure} as a building block
of a residual network \cite{He2016} with a residual function containing
only a non-linear transform and a shortcut connection containing an
affine transform.

The discretized equation (\ref{eq:numer-forward}) retains some of
the scaling properties of the differential equation (\ref{eq:main}).
Rescaled activations $\boldsymbol{x}_{\mathrm{s}}^{(l)}=c\boldsymbol{x}^{(l)}$,
where $c$ is some constant, lead to the same equation (\ref{eq:numer-forward})
only with scaled biases $\tilde{\boldsymbol{b}}_{\mathrm{s}}^{(l)}=c\tilde{\boldsymbol{b}}^{(l)}$
and scaled parameter of nonlinearity $\tilde{\gamma}_{\mathrm{s}}=c^{-2}\tilde{\gamma}$.
Thus we can set $\tilde{\gamma}_{\mathrm{s}}=1$ by taking $c=\sqrt{\tilde{\gamma}}$.

\subsection{Planar data classification}

\begin{figure}
\begin{centering}
\includegraphics[height=5.2cm]{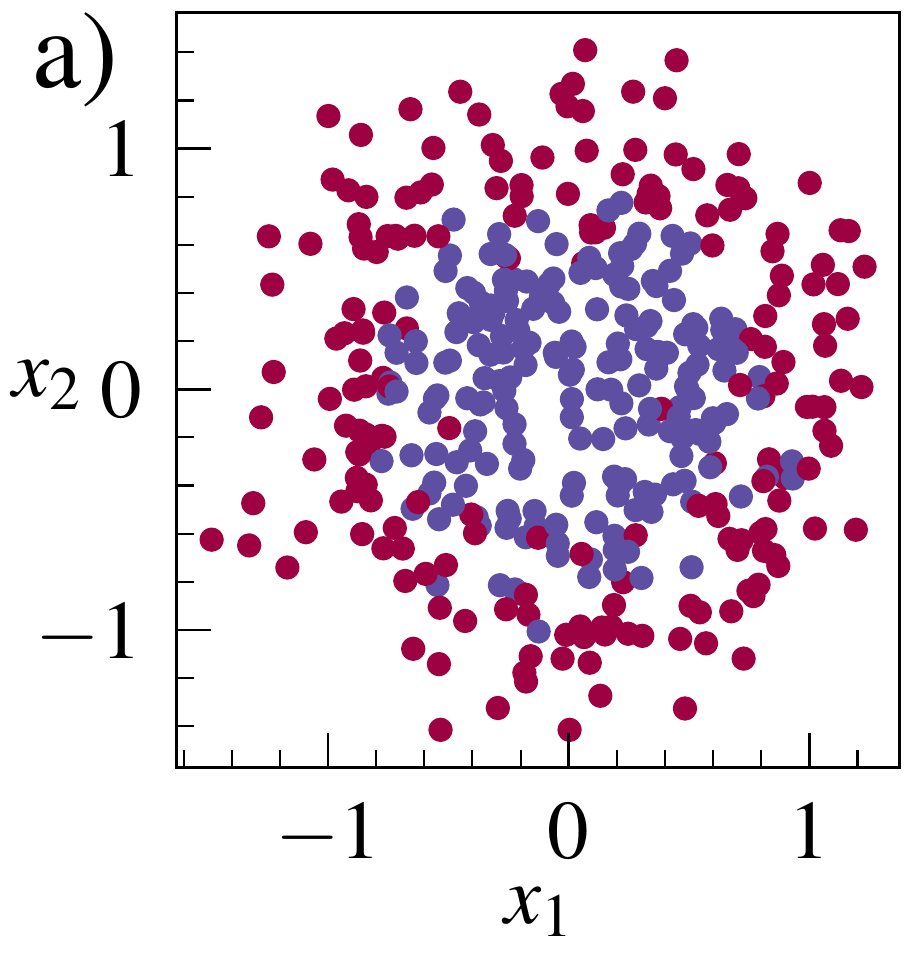}\hfill{}\includegraphics[height=5.2cm]{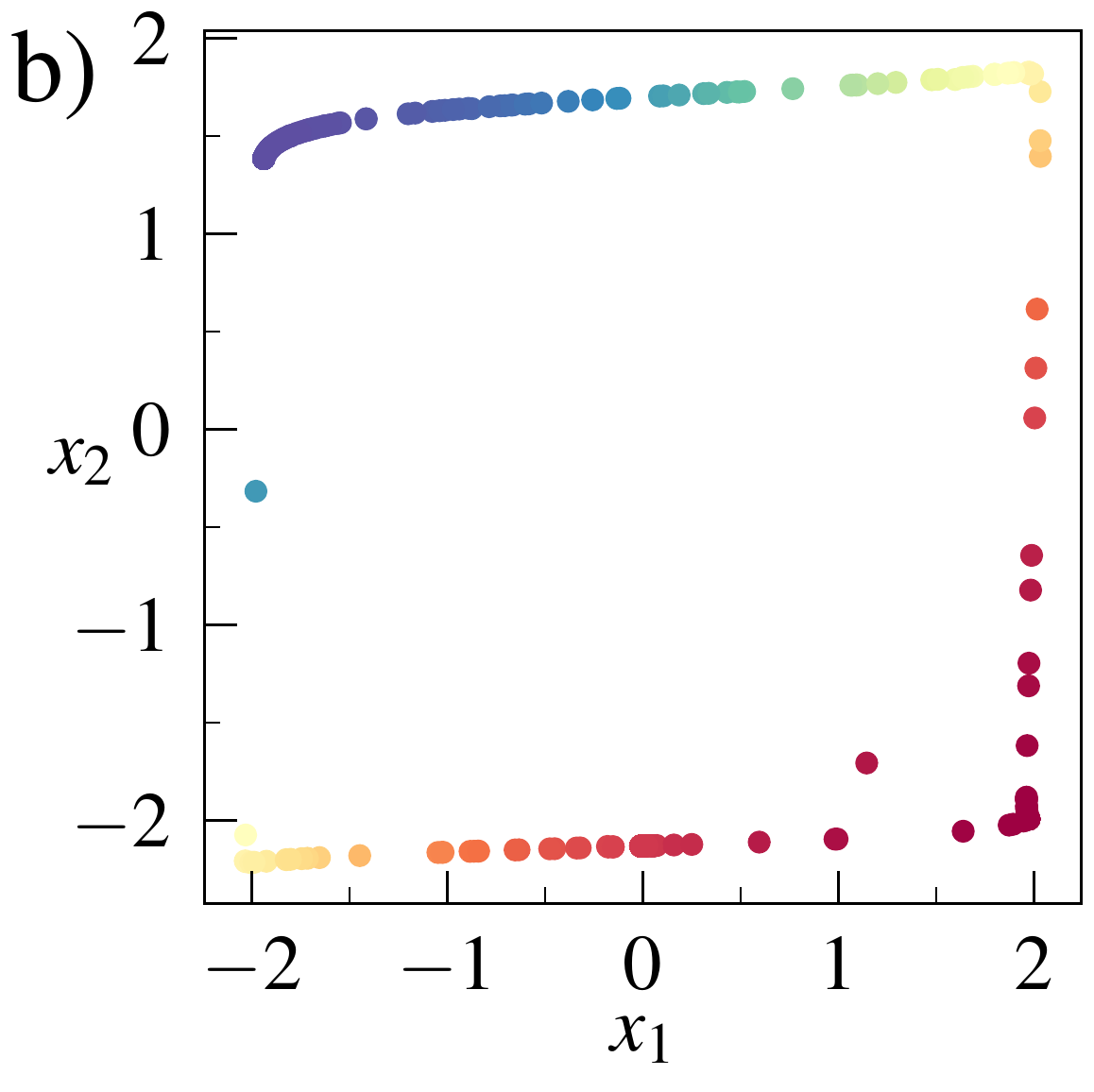}\hfill{}\includegraphics[height=5.2cm]{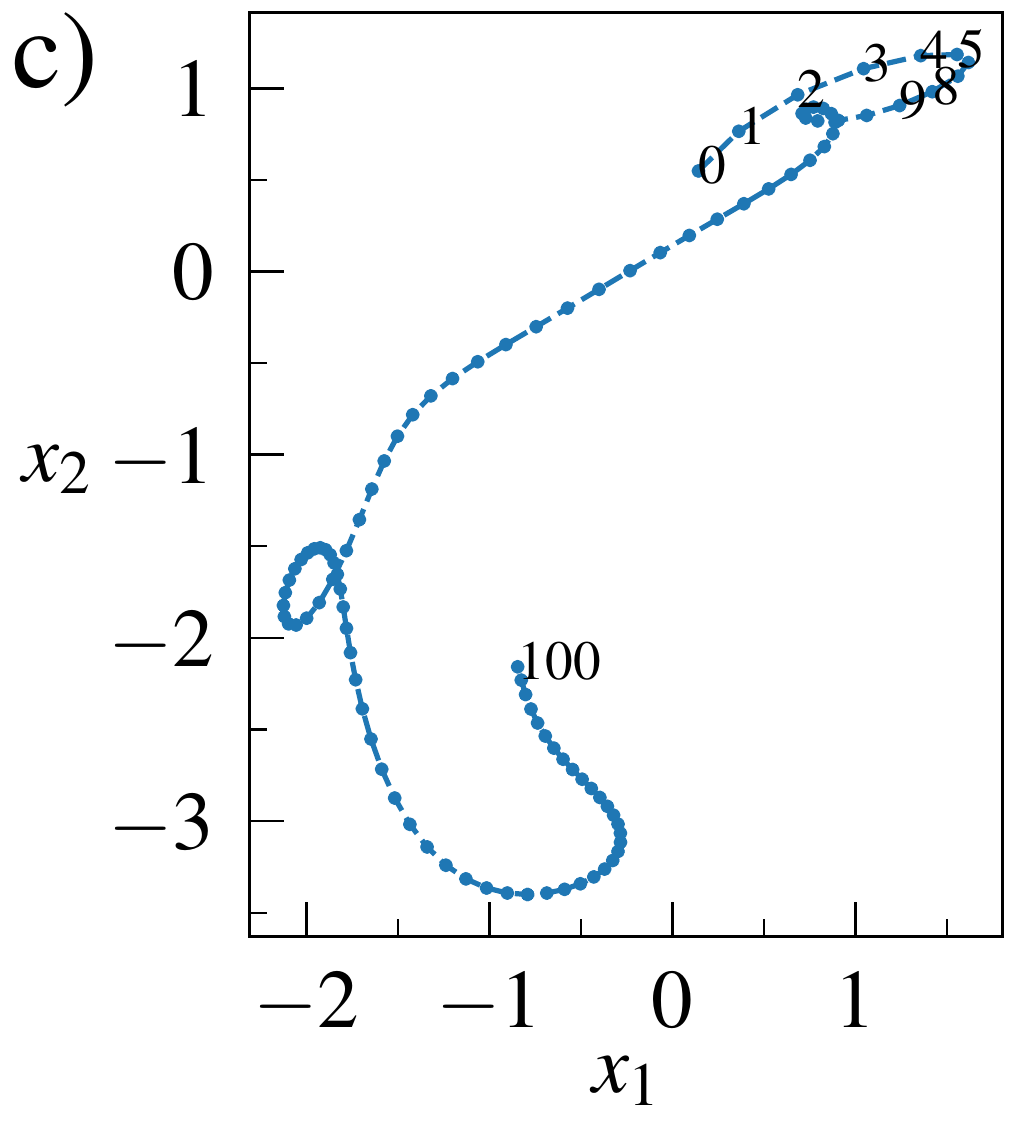}
\par\end{centering}
\caption{a) Dataset consisting of two concentric circles with an additional
Gaussian noise. Two classes $1$ and $2$ are shown with different
colors (blue and red). b) Transformed data in the output of the first
neural network, the color of a point indicates the probability to
be assigned a class $2$. c) Sample trajectory of an input produced
by the first neural network. Numbers beside the trajectory indicate
the layer number $l$.}
\label{fig:points-dataset}
\end{figure}

\begin{figure}
\centering{}\includegraphics[width=0.45\textwidth]{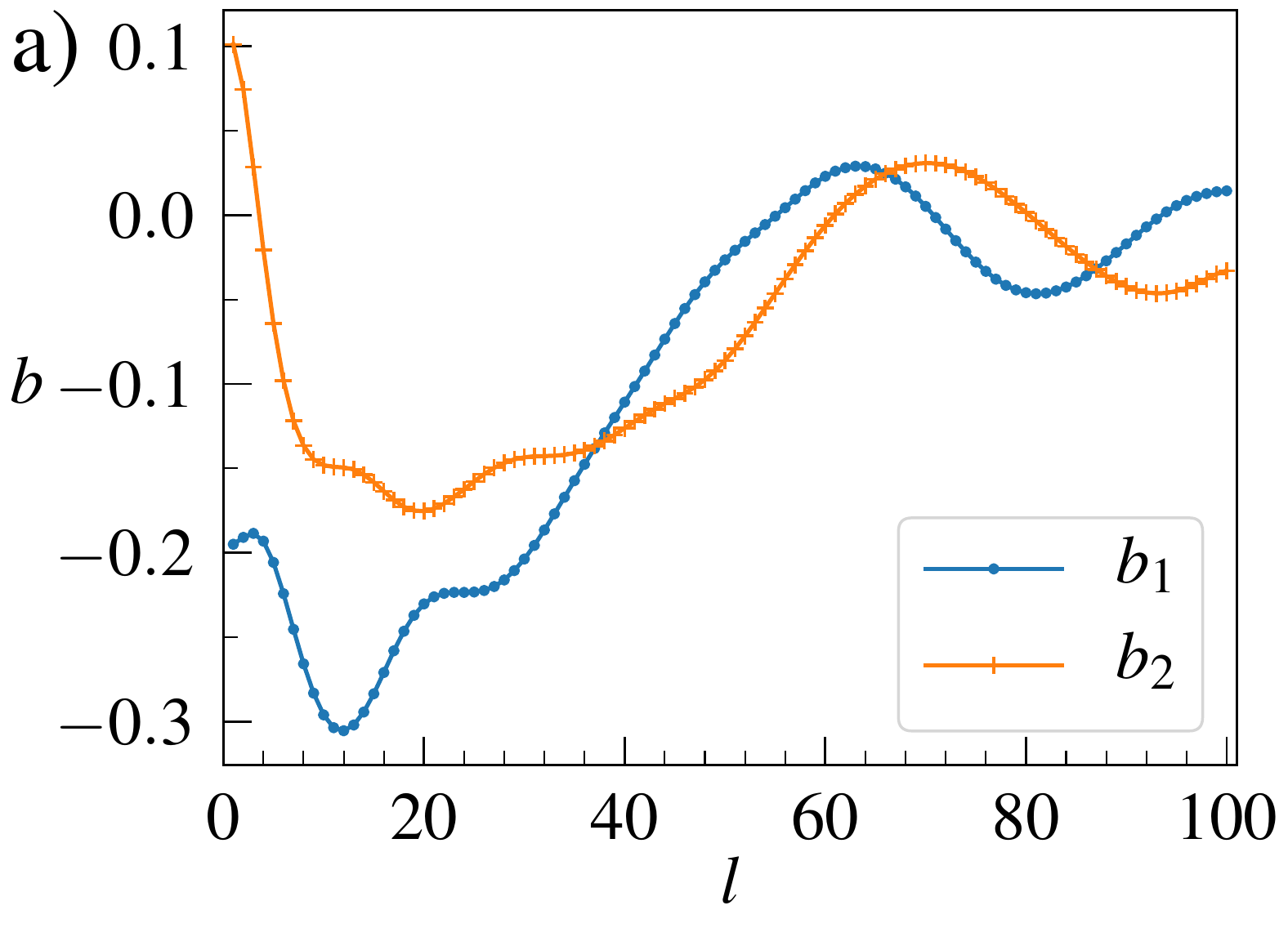}\hfill{}\includegraphics[width=0.45\textwidth]{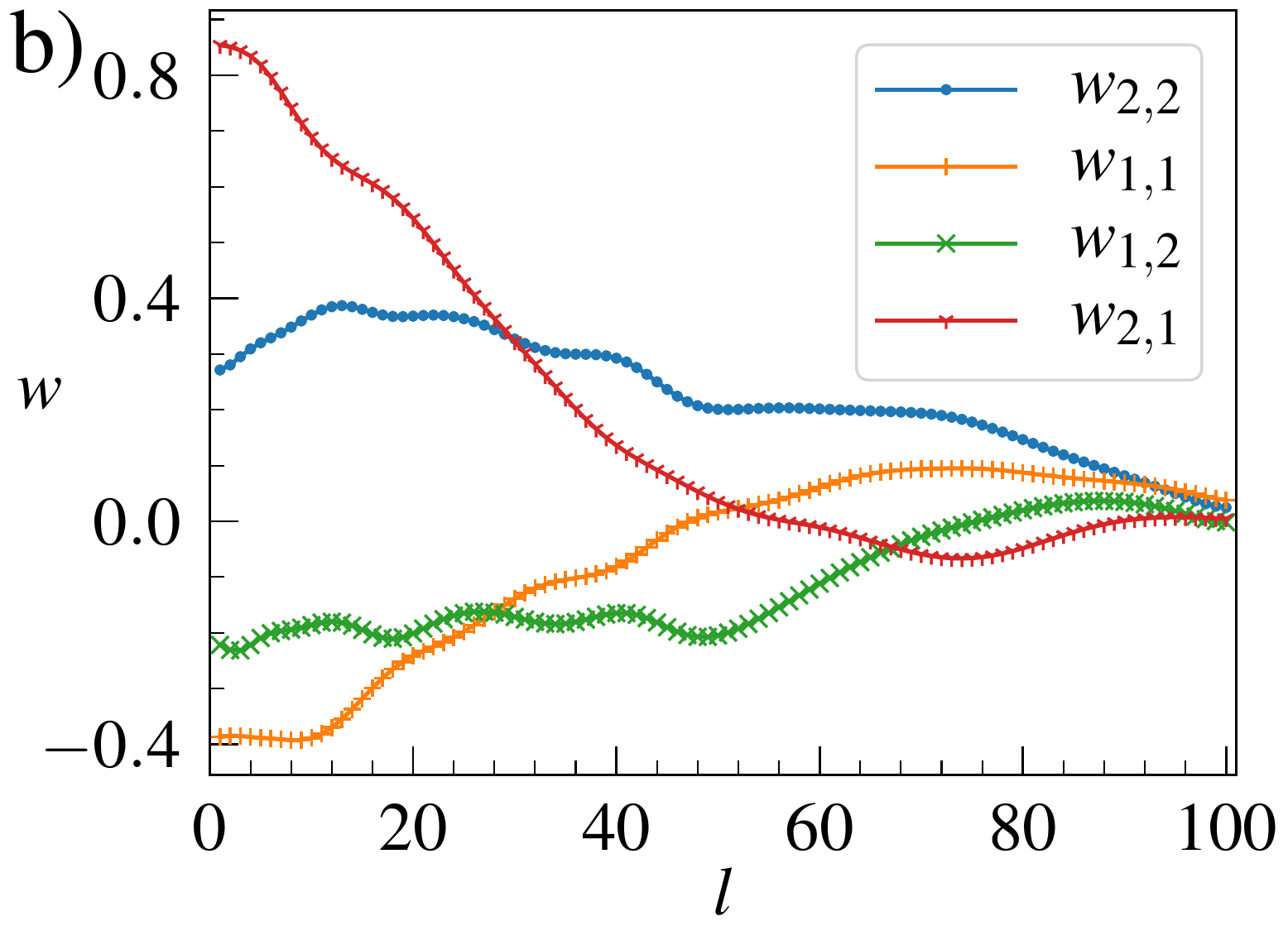}
\caption{Dependence of the parameters of the trained first model on the layer
number $l$: a) biases $b_{1}$ (circles) and $b_{2}$ (plus markers);
b) matrix elements of the weights matrix $\tilde{\boldsymbol{w}}$.}
\label{fig:points-parameters}
\end{figure}

In order to demonstrate the behavior of the model described by the
differential equation (\ref{eq:main}) it is convenient to consider
data described by only two features $x_{1}$ and $x_{2}$, because
such data can be visualized as points in the Euclidean plane. We limit
the number of classes to two and will consider a dataset shown in
figure \ref{fig:points-dataset} a). The dataset consist $400$ samples
of a Gaussian noise with standard deviation $0.2$ added to the two
concentric circles where the outer circle has two times larger radius
than the inner circle.

In the first model the differential equation (\ref{eq:main}) is solved
on the $\mathbb{R}^{2}$ space, giving the solutions $x_{1}(t)$,
$x_{2}(t)$. According to \cite{Dupont2019}, the dataset is hard
to learn using such an equation, because the differential equation
can only continuously deform the input space. Using the forward Euler
method, described by the equation (\ref{eq:numer-forward}), we discretize
the continuous variable $t$ into $L=100$ steps, producing a neural
network with $100$ layers, each layer containing $N=2$ units. Softmax
of the output of the final layer is calculated to get the predicted
probabilities of the two classes. Similar to learnable parameter of
nonlinearity in Parametric ReLUs \cite{He2015}, we consider the nonlinearity
$\tilde{\gamma}$ as an additional parameter to be learned. Thus the
neural network has $601$ trainable parameters. Employing the regularization
method described in section \ref{subsec:regularization} we minimize
the changes of the parameters between the neighboring layers and use
the loss function
\begin{equation}
\mathcal{L}=\mathcal{L}_{\mathrm{ce}}+\frac{\lambda}{2}\sum_{l}\left(\sum_{i}(\tilde{b}_{i}^{(l)}-
\tilde{b}_{i}^{(l-1)})^{2}+\sum_{i,j}(\tilde{w}_{i,j}^{(l)}-
\tilde{w}_{i,j}^{(l-1)})^{2}\right)\,,\label{eq:loss-smooth}
\end{equation}
where $\mathcal{L}_{\mathrm{ce}}$ is the usual cross entropy loss.
We take the regularization parameter $\lambda=1.0$ in the loss (\ref{eq:loss-smooth}).
All trainable parameters of the neural network are initialized to
zero, as described in section \ref{subsec:initialization}, and optimized
using the stochastic gradient descent. The parameters $\tilde{\boldsymbol{b}}$
and $\tilde{\boldsymbol{w}}$ of the trained model are shown in figure
\ref{fig:points-parameters} a) and b), respectively. We see that
indeed the loss (\ref{eq:loss-smooth}) produces smooth variations
of the parameters with the layer number $l$. The nonlinearity parameter
$\tilde{\gamma}$ acquires the value $\tilde{\gamma}=0.014$. Since
each layer of the network has two units, we can interpret all activations
as points in the same plane. After training, the transformed data
in the output of the network is shown in the figure \ref{fig:points-dataset}
b). As one can see, the points from the inner circle are moved to
the upper left part of the plane, the points from the outer circle
are moved to the lower right part. Plotting activations of each layer
in the same plane we can draw the gradual transformation of an input
by the differential equation (\ref{eq:main}) as a trajectory. Sample
trajectory is shown in the figure \ref{fig:points-dataset} c). The
trajectory has several parts that look like rotations of the point
in the $x_{1}$-$x_{2}$ plane, consistent with the description in
section \ref{subsec:weights}. However, the trajectory is complicated,
reflecting the difficulty to learn the dataset.

\begin{figure}
\centering{}\includegraphics[width=0.45\textwidth]{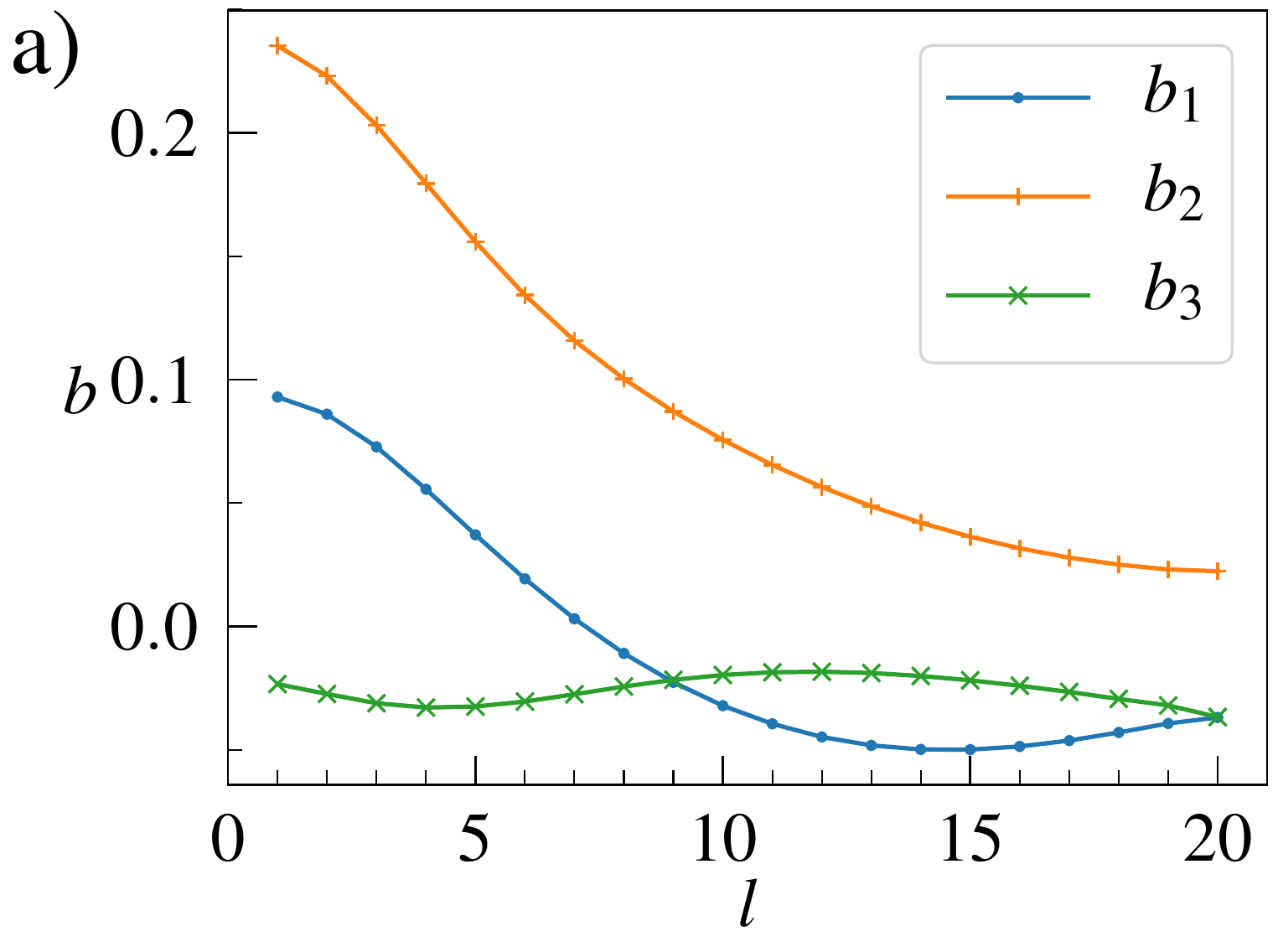}\hfill{}\includegraphics[width=0.45\textwidth]{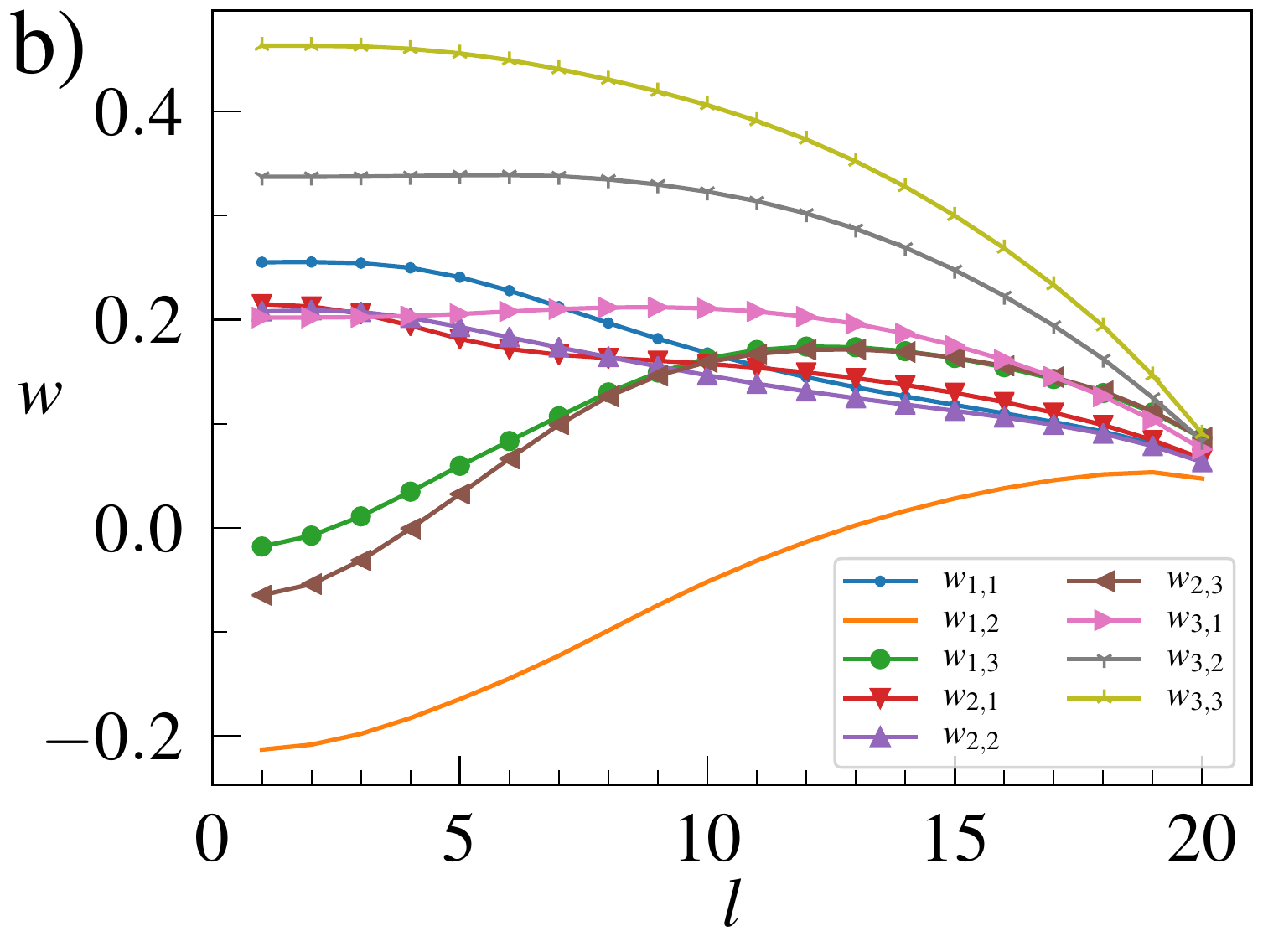}
\caption{Dependence of the parameters of the trained second model on the layer
number $l$: a) biases $\tilde{\boldsymbol{b}}$; b) matrix elements
of the weights matrix $\tilde{\boldsymbol{w}}$.}
\label{fig:points-3-parameters}
\end{figure}

\begin{figure}
\includegraphics[height=5.2cm]{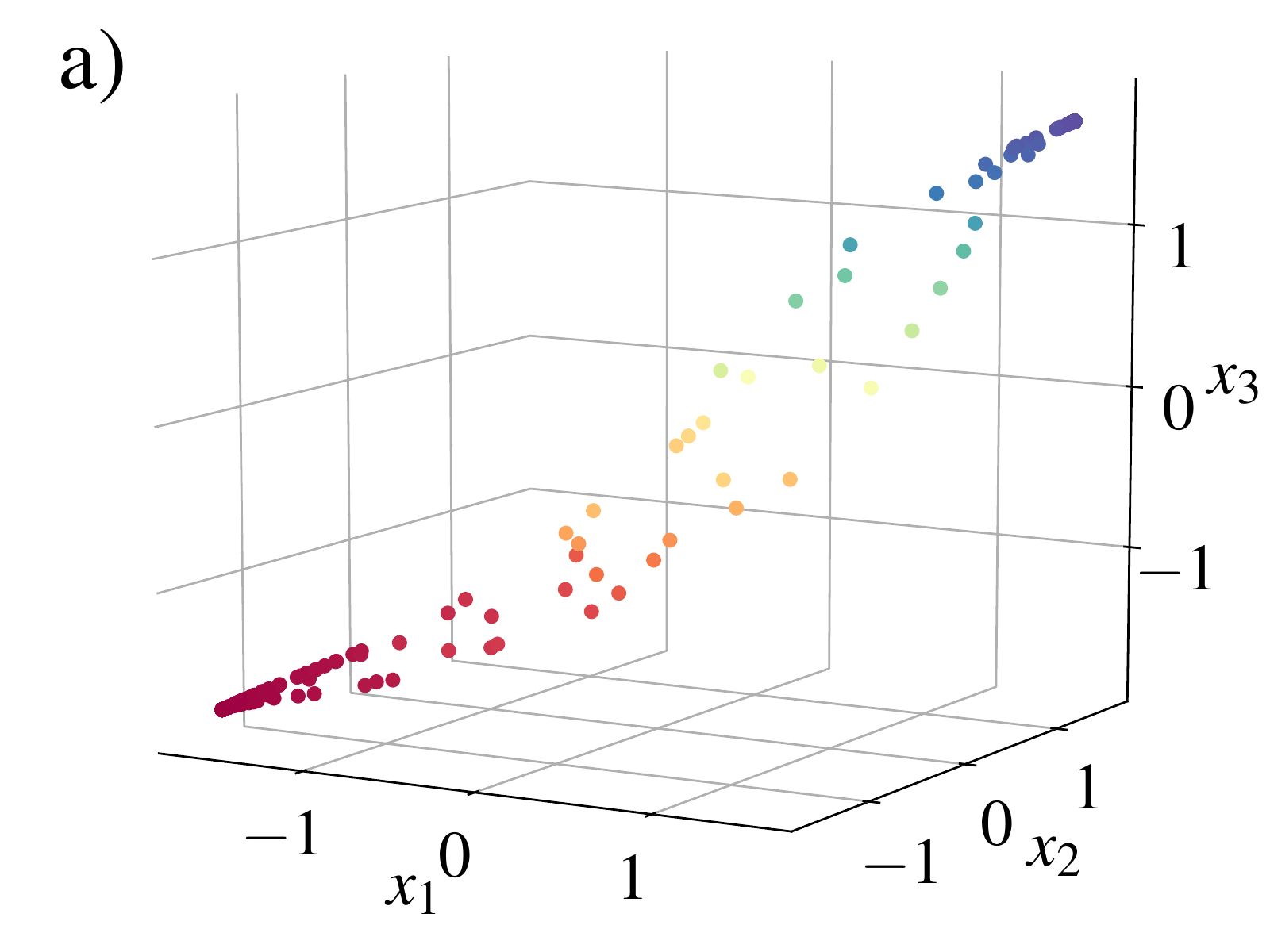}\hfill{}\includegraphics[height=5.2cm]{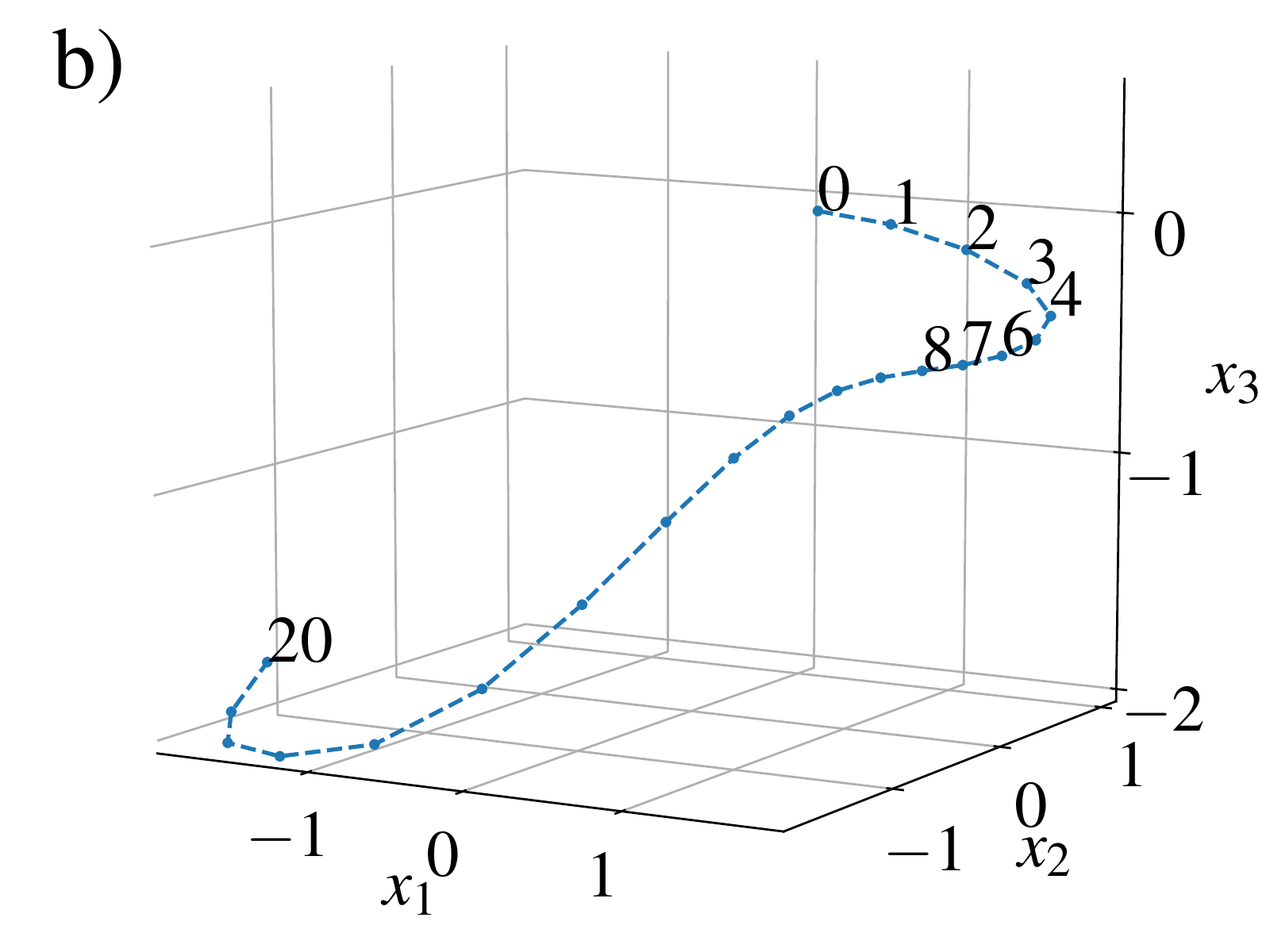}
\caption{a) Transformed data in the output of the second neural network, the
color of a point indicates the probability to be assigned a class
$2$. b) Sample trajectory of an input produced by the second neural
network. Numbers beside the trajectory indicate the layer number $l$.}
\label{fig:points-3-dataset}
\end{figure}

In the second model we solve the differential equation on the extended
$\mathbb{R}^{3}$ space, taking the initial condition $x_{3}(0)=0$.
We discretize the continuous variable $t$ into $L=20$ steps, producing
a neural network with $20$ layers, each layer containing $N=3$ units.
The predicted probability of the first class is calculated as sigmoid
of $x_{3}^{(L)}$. We use loss (\ref{eq:loss-smooth}) with the regularization
parameter $\lambda=1.0$ and fixed nonlinearity parameter $\tilde{\gamma}=0.1$.
Consequently, the neural network has $240$ trainable parameters.
The parameters $\tilde{\boldsymbol{b}}$ and $\tilde{\boldsymbol{w}}$
of the trained model are shown in figure \ref{fig:points-3-parameters}
a) and b), respectively. After training, the transformed data in the
output of the network is shown in the figure \ref{fig:points-3-dataset}
a). As one can see, the points from the inner circle are moved to
the upper part $x_{3}>0$, the points from the outer circle are moved
to the lower part $x_{3}<0$. Sample trajectory, showing the gradual
transformation of an input by the differential equation (\ref{eq:main})
is shown in the figure \ref{fig:points-3-dataset} b). We see that,
compared with the first model, the trajectory is much simpler, indicating
that the dataset is easier to learn using the second model. This fact
that wider models are superior to thin and deep counterparts has been
noticed for ResNets \cite{Zagoruyko2016}.

\subsection{MNIST digit recognition}

\begin{figure}
\includegraphics[width=0.45\textwidth]{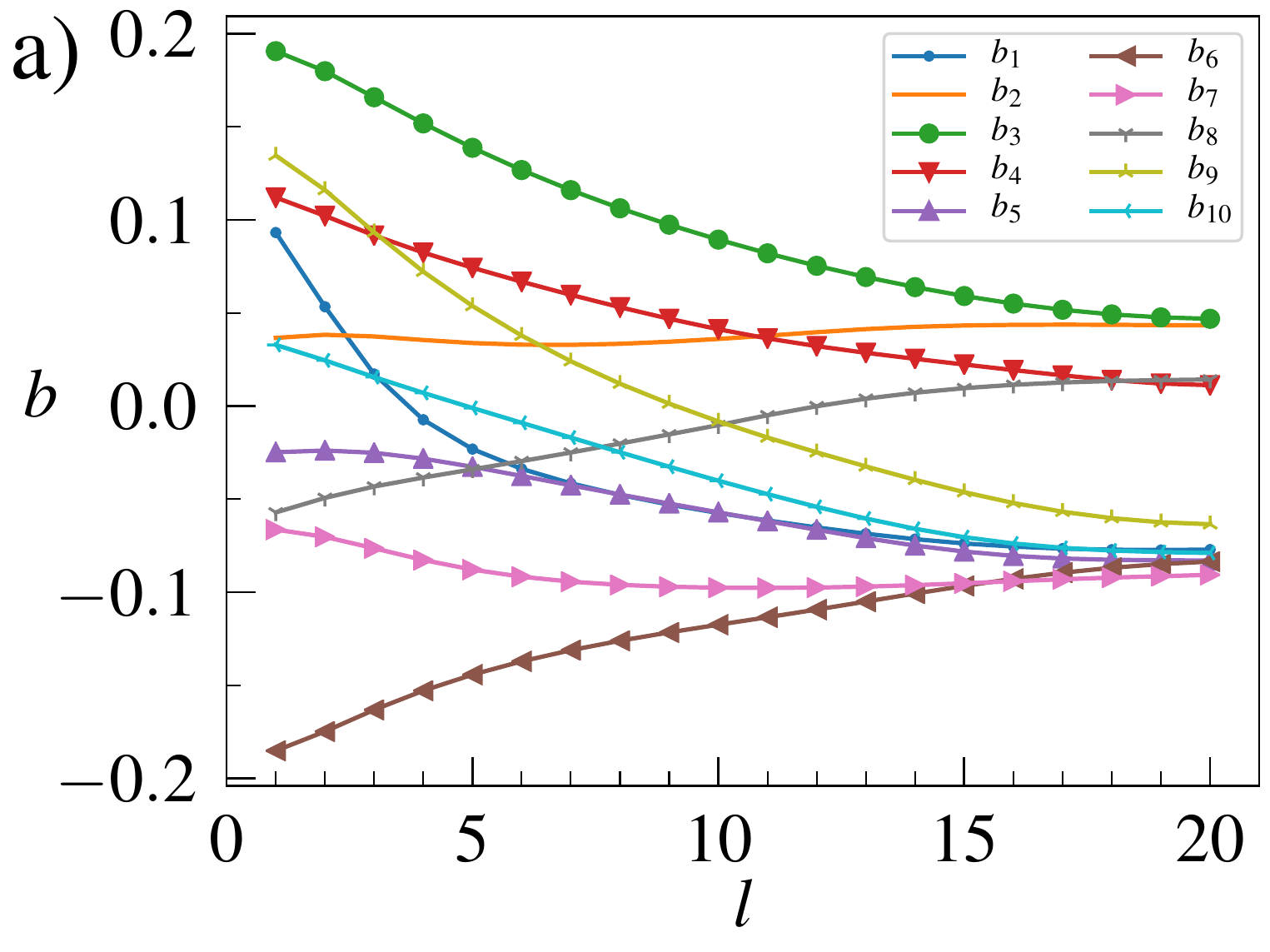}\hfill{}\includegraphics[width=0.45\textwidth]{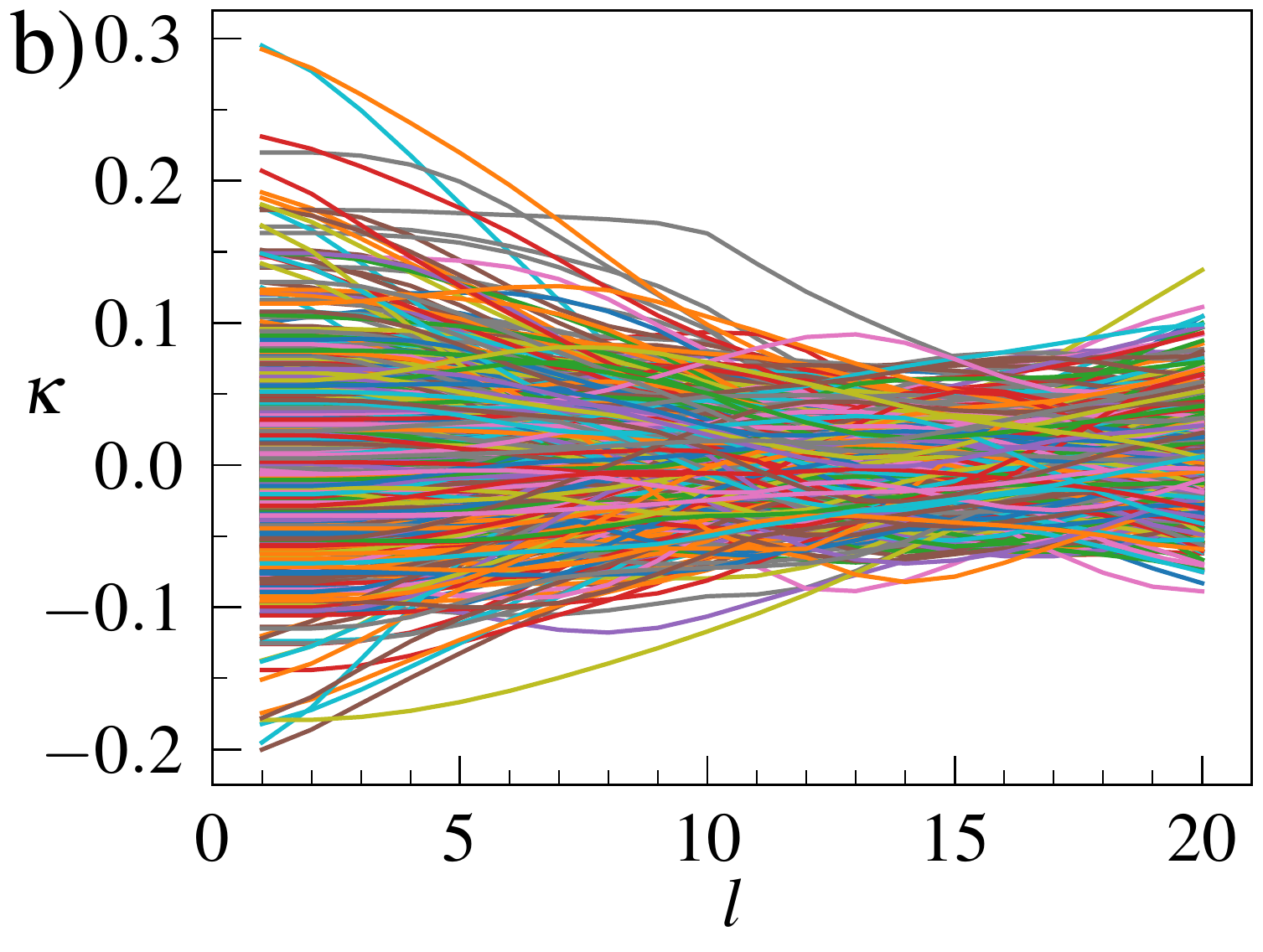}
\caption{Dependence of the parameters of the trained convolutional model on
the layer number $l$: a) biases $\tilde{\boldsymbol{b}}$; b) elements
of the convolution kernel $\tilde{\boldsymbol{\kappa}}$.}
\label{fig:mnist-parameters}
\end{figure}

\begin{figure}
\parbox[t]{0.55\textwidth}{%
\vspace{0pt}
\includegraphics[width=0.55\textwidth]{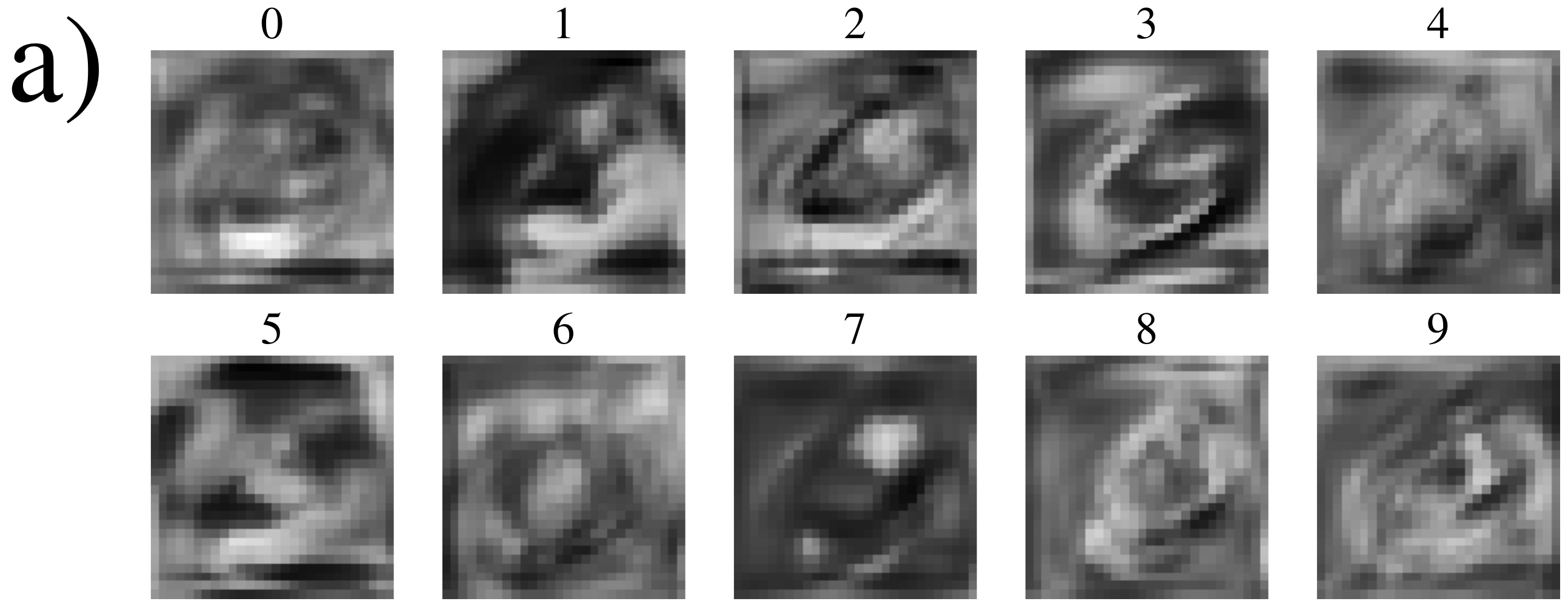}%
}\hfill{}%
\parbox[t]{0.4\textwidth}{%
\vspace{0pt}
\includegraphics[width=0.4\textwidth]{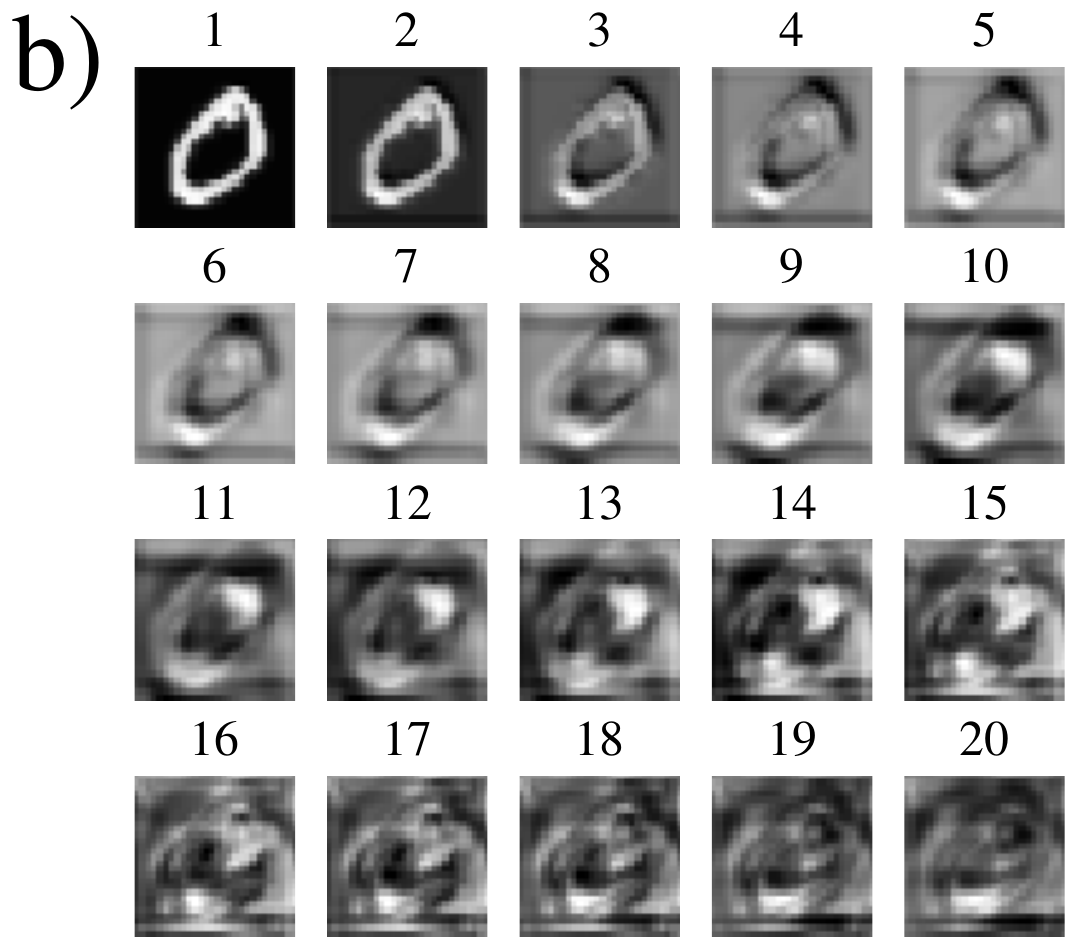}%
}
\caption{a) Sample activations of the final layer of the network when the initial
image contains digit ``0''. b) Corresponding activations in the first
channel after each layer.}
\label{fig:mnist-trajectory}
\end{figure}

To demonstrate the behavior of the convolutional model described by
the differential equation (\ref{eq:conv-forward}) we will consider
recognition of digits from MNIST dataset. As is suggested in \cite{Dupont2019},
we augment the solution space by adding new channels of zeros to the
input image. We choose the number of additional channels equal to
$9$ thus making the total number of channels equal to $10$, which
also is the number of classes. In the model we use convolutions with
$3\times3$ filter size. We discretize the continuous variable $t$
into $L=20$ steps, producing a neural network with $20$ convolutional
layers. After the last convolutional layer we perform global max pooling
and take softmax of the resulting $10$ numbers as the predicted probabilities
of the $10$ classes. We consider the nonlinearity $\tilde{\gamma}$
as an additional parameter to be learned. The resulting neural network
has $(3\times3\times10\times10+10)\times20+1=18201$ trainable parameters.
As in previous subsection, we use the regularization method that minimize
the changes of the parameters between the neighboring layers, taking
the regularization parameter $\lambda=0.1$. All trainable parameters
of the neural network are initialized to zero, as described in section
\ref{subsec:initialization}, and optimized using the stochastic gradient
descent. After training the network achieves 97.8\% accuracy. The
nonlinearity parameter $\tilde{\gamma}$ acquires the value
$\tilde{\gamma}=4.8\cdot10^{-4}$.

The parameters $\tilde{\boldsymbol{b}}$ and $\tilde{\boldsymbol{\kappa}}$
of the trained model are shown in figure \ref{fig:mnist-parameters}
a) and b), respectively. As in the previous subsection we see smooth
variations of the parameters with the layer number $l$. Sample activations
of the final layer of the network when the initial image contains
digit ``0'' are shown in the figure \ref{fig:mnist-trajectory} a)
and the corresponding activations in the first channel after each
layer are shown in figure \ref{fig:mnist-trajectory} b). As we can
see in figure \ref{fig:mnist-trajectory} b), the input image is gradually
transformed by the neural network to the output activations.

\section{Conclusions\label{sec:conclusions}}

To summarize, we investigated nonlinear differential equations (\ref{eq:diff-general})
as machine learning models. The gradient of the loss function with
respect to to the hidden state can be considered as a generalized
momentum conjugate to the hidden state, and the differential equation
for forward and backward propagation can be recast in Euler-Lagrange
or Hamilton form. The corresponding Lagrangian and Hamiltonian are
given by equations (\ref{eq:lagrangian}) and (\ref{eq:hmiltonian}).
The Euler-Lagrange and the Hamilton form of the equations allows analyze
consequences of the possible symmetries in the models. Furthermore,
we showed that not only residual networks, but also feedforward neural
networks with small nonlinearities and the weights matrices deviating
only slightly from identity matrices can be related to the differential
equations. We proposed a differential equation (\ref{eq:main}) describing
such networks.

%\bibliographystyle{unsrt}
%\bibliography{nn}

\begin{thebibliography}{10}

\bibitem{Bengio2009}
Y.~Bengio.
\newblock Learning deep architectures for {AI}.
\newblock {\em Found. Trends Mach. Learn.}, 2(1):1--127, 2009.

\bibitem{LeCun2015}
Y.~LeCun, Y.~Bengio, and G.~Hinton.
\newblock Deep learning.
\newblock {\em Nature}, 521:436, 2015.

\bibitem{Sonoda2017}
S.~Sonoda and N.~Murata.
\newblock Double continuum limit of deep neural networks.
\newblock In {\em ICML 2017 Workshop on Principled Approaches to Deep
  Learning}, 2017.

\bibitem{Grossberg2013}
S.~Grossberg.
\newblock Recurrent neural networks.
\newblock {\em Scholarpedia}, 8(2):1888, 2013.
\newblock revision \#138057.

\bibitem{Cohen1983}
M.~A. Cohen and S.~Grossberg.
\newblock Absolute stability of global pattern formation and parallel memory
  storage by competitive neural networks.
\newblock {\em IEEE Transactions on Systems, Man, and Cybernetics},
  SMC-13(5):815--826, 1983.

\bibitem{Hopfield1984}
J.~J. Hopfield.
\newblock Neurons with graded response have collective computational properties
  like those of two-state neurons.
\newblock {\em Proc. Natl. Acad. Sci. USA}, 81:3088--3092, 1984.

\bibitem{Hopfield2007}
J.~J. Hopfield.
\newblock {H}opfield network.
\newblock {\em Scholarpedia}, 2(5):1977, 2007.
\newblock revision \#91363.

\bibitem{Harvey1994}
I.~Harvey, P.~Husbands, and D.~Cliff.
\newblock Seeing the light: Artificial evolution, real vision.
\newblock In {\em 3rd international conference on Simulation of adaptive
  behavior: from animals to animats}, pages 392--401, 1994.

\bibitem{Beer1997}
R.~D. Beer.
\newblock The dynamics of adaptive behavior: A research program.
\newblock {\em Robotics and Autonomous Systems}, 20(2-4):257--289, 1997.

\bibitem{Quinn2001}
Matt Quinn.
\newblock Evolving communication without dedicated communication channels.
\newblock In Jozef Kelemen and Petr Sos{\'i}k, editors, {\em Advances in
  Artificial Life}, pages 357--366. Springer, 2001.

\bibitem{He2016}
K.~He, X.~Zhang, S.~Ren, and J.~Sun.
\newblock Deep residual learning for image recognition.
\newblock In {\em 2016 IEEE Conference on Computer Vision and Pattern
  Recognition (CVPR)}, pages 770--778, June 2016.

\bibitem{Liao2016}
Q.~Liao and T.~Poggio.
\newblock Bridging the gaps between residual learning, recurrent neural
  networks and visual cortex.
\newblock arXiv:1604.03640, 2016.

\bibitem{E2017}
W.~E.
\newblock A proposal on machine learning via dynamical systems.
\newblock {\em Commun. Math. Stat.}, 5:1--11, 2017.

\bibitem{Haber2017}
E.~Haber, L.~Ruthotto, E.~Holtham, and S.-H. Jun.
\newblock Learning across scales -- multiscale methods for convolution neural
  networks.
\newblock arXiv:1703:02009, 2017.

\bibitem{Haber2018}
E.~Haber and L.~Ruthotto.
\newblock Stable architectures for deep neural networks.
\newblock {\em Inverse Problems}, 34:014004, 2018.

\bibitem{Chang2017}
B.~Chang, L.~Meng, E.~Haber, L.~Ruthotto, D.~Begert, and E.~Holtham.
\newblock Reversible architectures for arbitrarily deep residual neural
  networks.
\newblock arXiv:1709.03698, 2017.

\bibitem{Behrmann2018}
J.~Behrmann, W.~Grathwohl, R.~T.~Q. Chen, D.~Duvenaud, and J.-H. Jacobsen.
\newblock Invertible residual networks.
\newblock arXiv:1811.00995, 2018.

\bibitem{Li2017}
Z.~Li and Z.~Shi.
\newblock Deep residual learning and pdes on manifold.
\newblock arXiv:1708.05115, 2017.

\bibitem{Lu2017}
Y.~Lu, A.~Zhong, Q.~Li, and B.~Dong.
\newblock Beyond finite layer neural networks: Bridging deep architectures and
  numerical differential equations.
\newblock arXiv:1710:10121, 2017.

\bibitem{Ruthotto2018}
L.~Ruthotto and E.~Haber.
\newblock Deep neural networks motivated by partial differential equations.
\newblock arXiv:1804.04272, 2018.

\bibitem{Long2017}
Z.~Long, Y.~Lu, X.~Ma, and B.~Dong.
\newblock {PDE}-{N}et: Learning {PDE}s from data.
\newblock arXiv:1710.09668, 2017.

\bibitem{Long2018}
Z.~Long, Y.~Lu, and B.~Dong.
\newblock {PDF}-{N}et 2.0: Learning {PDE}s from data with a numeric-symbolic
  hybrid deep network.
\newblock arXiv:1812:04426, 2018.

\bibitem{Niu2019}
M.~Y. Niu, I.~L. Chuang, and L.~Horesh.
\newblock Recurrent neural networks in the eye of differential equations.
\newblock arXiv:1904.12933, 2019.

\bibitem{Chen2018}
T.~Q. Chen, Y.~Rubanova, J.~Bettencourt, and D.~K. Duvenaud.
\newblock Neural ordinary differential equations.
\newblock In S.~Bengio, H.~Wallach, H.~Larochelle, K.~Grauman, N.~Cesa-Bianchi,
  and R.~Garnett, editors, {\em Advances in Neural Information Processing
  Systems 31}, pages 6571--6583. Curran Associates, Inc., 2018.

\bibitem{Rubanova2019}
Y.~Rubanova, R.~T.~Q. Chen, and D.~Duvenaud.
\newblock Latent {ODEs} for irregularly-sampled time series.
\newblock arXiv:1907.03907, 2019.

\bibitem{Dupont2019}
E.~Dupont, A.~Doucet, and Y.~W. Teh.
\newblock Augmented neural {ODE}s.
\newblock arXiv:1904:01681, 2019.

\bibitem{Pontryagin2018}
L.~S. Pontryagin, E.~F. Mishchenko, V.~G. Boltyanskii, and R.~V. Gamkrelidze.
\newblock {\em The mathematical theory of optimal processes}.
\newblock Classics of Soviet mathematics. Routledge, Boca Raton, 4th ed
  edition, 2018.

\bibitem{Giaquinta2006}
M.~Giaquinta and S.~Hildebrandt.
\newblock {\em Calculus of variations I}.
\newblock Grundlehren der mathematischen Wissenschaften Vol 1. Springer, 2006.

\bibitem{Rumelhart1986}
D.~E. Rumelhart, G.~E. Hinton, and R.~J. Williams.
\newblock Learning representations by back-propagating errors.
\newblock {\em Nature}, 323:533--536, 1986.

\bibitem{Ibragimov2006}
N.~H. Ibragimov.
\newblock Integrating factors, adjoint equations and lagrangians.
\newblock {\em J. Math. Anal. Appl.}, 318:742--757, 2006.

\bibitem{Noether1918}
E.~Noether.
\newblock Invariante variationsprobleme.
\newblock In {\em G{\"o}ttingen Math. Phys. Kl.}, pages 235--257.
  K{\"o}nigliche Gesellschaft der Wissenschaften, 1918.
\newblock English transl.: Transport Theory and Statistical Physics 1(3):
  186--207, 1971.

\bibitem{Srivastava2015}
R.~K. Srivastava, K.~Greff, and J.~Schmidhuber.
\newblock Highway networks.
\newblock arXiv:1505.00387, 2015.

\bibitem{Srivastava2015a}
R.~K. Srivastava, K.~Greff, and J.~Schmidhuber.
\newblock Training very deep networks.
\newblock arXiv:1507.06228, 2015.

\bibitem{He2015}
K.~He, X.~Zhang, S.~Ren, and J.~Sun.
\newblock Delving deep into rectifiers: Surpassing human-level performance on
  imagenet classification.
\newblock In {\em 2015 IEEE International Conference on Computer Vision
  (ICCV)}, pages 1026--1034, Dec 2015.

\bibitem{Hinton2012}
G.~E. Hinton, N.~Srivastava, A.~Krizhevsky, I.~Sutskever, and R.~R.
  Salakhutdinov.
\newblock Improving neural networks by preventing co-adaptation of feature
  detectors.
\newblock arXiv:1207.0580, 2012.

\bibitem{Simard2003}
P.~Y. Simard, D.~Steinkraus, and J.~Platt.
\newblock Best practices for convolutional neural networks applied to visual
  document analysis.
\newblock In {\em Seventh International Conference on Document Analysis and
  Recognition, 2003. Proceedings.}, pages 958--963. Institute of Electrical and
  Electronics Engineers, Inc., August 2003.

\bibitem{Montavon2012}
G.~Montavon, G.~Orr, and K.-R. M{\"u}ller, editors.
\newblock {\em Neural networks: tricks of the trade}.
\newblock Springer, 2012.

\bibitem{Morgan1990}
N.~Morgan and H.~Bourlard.
\newblock Generalization and parameter estimation in feedforward nets: Some
  experiments.
\newblock In D.~S. Touretzky, editor, {\em Advances in Neural Information
  Processing Systems 2}, pages 630--637. Morgan Kaufmann Publishers Inc., San
  Francisco, CA, USA, 1990.

\bibitem{Srivastava2014}
N.~Srivastava, G.~E. Hinton, A.~Krizhevsky, I.~Sutskever, and R.~Salakhutdinov.
\newblock Dropout: a simple way to prevent neural networks from overfitting.
\newblock {\em Journal of Machine Learning Research}, 15(1):1929--1958, 2014.

\bibitem{Glorot2010}
X.~Glorot and Y.~Bengio.
\newblock Understanding the difficulty of training deep feedforward neural
  networks.
\newblock In Y.~W. Teh and M.~Titterington, editors, {\em Proceedings of the
  Thirteenth International Conference on Artificial Intelligence and
  Statistics}, volume~9 of {\em Proceedings of Machine Learning Research},
  pages 249--256, Chia Laguna Resort, Sardinia, Italy, 13--15 May 2010. PMLR.

\bibitem{Krizhevsky2012}
A.~Krizhevsky, I.~Sutskever, and G.~E. Hinton.
\newblock Imagenet classification with deep convolutional neural networks.
\newblock In {\em Proceedings of the 25th International Conference on Neural
  Information Processing Systems - Volume 1}, NIPS'12, pages 1097--1105, USA,
  2012. Curran Associates Inc.

\bibitem{Butcher2003}
J.~C. Butcher.
\newblock {\em Numerical Methods for Ordinary Differential Equations}.
\newblock J. Wiley, 2003.

\bibitem{Zagoruyko2016}
S.~Zagoruyko and N.~Komodakis.
\newblock Wide residual networks.
\newblock In {\em BMVC}, 2016.

\end{thebibliography}

\appendix
\numberwithin{equation}{section}

\section{Alternative method of derivation of differential equation for backward
propagation \label{sec:appendix}}

Instead of variational calculus we can employ a simpler method by
discretizing the equation (\ref{eq:diff-general}) using a small time
step $\Delta t$ as
\begin{equation}
\frac{1}{\Delta t}[\boldsymbol{x}(t_{l+1})-\boldsymbol{x}(t_{l})]=
\boldsymbol{F}(\boldsymbol{x}(t_{l}),\boldsymbol{q}(t_{l}))\label{eq:diff-discr}
\end{equation}
with $t_{l}=l\Delta t$ and taking the limit $\Delta t\rightarrow0$
afterwards. This equation leads to the following Jacobian of the layer
transform:
\begin{equation}
\frac{\partial\boldsymbol{x}(t_{l+1})}{\partial\boldsymbol{x}(t_{l})}=
\boldsymbol{I}+\Delta t\frac{\partial}{\partial\boldsymbol{x}(t_{l})}
\boldsymbol{F}(\boldsymbol{x}(t_{l}),\boldsymbol{q}(t_{l}))\,.\label{eq:jacobian}
\end{equation}
In order to calculate the gradients of a loss function $\mathcal{L}$
with respect to the parameters of the network, one needs to have the
gradients with respect to the activations $\boldsymbol{x}^{(l)}$
of each layer. The gradient of the loss with respect to the activations
of the $l$-th layer is related to the gradient with respect to the
activations of the $l+1$-th layer via the equation
\begin{equation}
\frac{\partial\mathcal{L}}{\partial\boldsymbol{x}(t_{l})}=
\frac{\partial\mathcal{L}}{\partial\boldsymbol{x}(t_{l+1})}
\frac{\partial\boldsymbol{x}(t_{l+1})}{\partial\boldsymbol{x}(t_{l})}\,.
\end{equation}
Inserting the Jacobian (\ref{eq:jacobian}) we obtain
\begin{equation}
\frac{1}{\Delta t}[\boldsymbol{a}(t_{l})-\boldsymbol{a}(t_{l+1})]=
\boldsymbol{a}(t_{l+1})\frac{\partial}{\partial\boldsymbol{x}(t_{l})}
\boldsymbol{F}(\boldsymbol{x}(t_{l}),\boldsymbol{q}(t_{l}))\,,
\end{equation}
where for the brevity we introduced the notation
\begin{equation}
\boldsymbol{a}(t_{l})\equiv\frac{\partial\mathcal{L}}{\partial\boldsymbol{x}(t_{l})}\,.
\end{equation}
Interpreting $t$ as a continuous variable and taking the limit $\Delta t\rightarrow0$
we get the differential equation (\ref{eq:backward-general}) for
the backward propagation.

The gradients of the loss function $\mathcal{L}$ with respect to
the parameters can be calculated using the equations
\begin{equation}
\frac{\partial\mathcal{L}}{\partial\boldsymbol{q}(t_{l})}=
\frac{\partial\mathcal{L}}{\partial\boldsymbol{x}(t_{l+1})}
\frac{\partial\boldsymbol{x}(t_{l+1})}{\partial\boldsymbol{q}(t_{l})}\label{eq:deriv-1}
\end{equation}
where the derivatives
\begin{equation}
\frac{\partial\boldsymbol{x}(t_{l+1})}{\partial\boldsymbol{q}(t_{l})}=
\Delta t\frac{\partial}{\partial\boldsymbol{q}(t_{l})}
\boldsymbol{F}(\boldsymbol{x}(t_{l}),\boldsymbol{q}(t_{l}))\label{eq:deriv-2}
\end{equation}
are obtained from the equation (\ref{eq:diff-discr}). Equations (\ref{eq:deriv-1}),
(\ref{eq:deriv-2}) lead to the gradients of the loss function
\begin{equation}
\frac{\partial\mathcal{L}}{\partial\boldsymbol{q}(t_{l})}=
\Delta t\boldsymbol{a}(t_{l+1})\frac{\partial}{\partial\boldsymbol{q}(t_{l})}
\boldsymbol{F}(\boldsymbol{x}(t_{l}),\boldsymbol{q}(t_{l}))\label{eq:deriv-3}
\end{equation}
In the limit $\Delta t\rightarrow0$ the gradients vanish. However,
taking the learning rate $\epsilon$ proportional to $\Delta t^{-1}$,
$\epsilon=\alpha\Delta t^{-1}$, one can get finite updates of the
parameters in the gradient descent step. If the functions $\boldsymbol{q}(t)$
depend on parameters $\boldsymbol{\theta}$ that do not depend on
$t$, $\boldsymbol{q}(t)=\boldsymbol{q}(\boldsymbol{\theta},t)$,
then the gradient of the loss can be calculated as
\begin{equation}
\frac{\partial\mathcal{L}}{\partial\boldsymbol{\theta}}=
\sum_{l}\frac{\partial\mathcal{L}}{\partial\boldsymbol{q}(t_{l})}
\frac{\partial\boldsymbol{q}(t_{l})}{\partial\boldsymbol{\theta}}\,.
\end{equation}
Using equation (\ref{eq:deriv-3}) and taking the limit $\Delta t\rightarrow0$
we get equation (\ref{eq:gradient-param}).
\end{document}